\documentclass[sigconf,nonacm]{acmart}

\setcopyright{none}
\ifdefined\XeTeXversion
  \usepackage{newunicodechar}
  \newunicodechar{–}{--}
\fi

\ifdefined\symbfit
  \newcommand{\bm}[1]{\symbfit{#1}}
\else
  \usepackage{bm}
\fi
\usepackage{multirow}
\usepackage{pgfplots,tikz}
\pgfplotsset{compat=1.18}
\usepgfplotslibrary{groupplots}
\usetikzlibrary{arrows,shapes}
\usetikzlibrary{automata,positioning}
\usetikzlibrary{calc}
\definecolor{darkgreen100}{RGB}{0,89,41}
\definecolor{darkgreen75}{RGB}{64,131,95}
\definecolor{darkgreen50}{RGB}{128,172,148}
\definecolor{darkgreen25}{RGB}{191,214,202}
\definecolor{darkgreen10}{RGB}{230,238,234}

\definecolor{brightgreen100}{RGB}{31,168,36}
\definecolor{brightgreen75}{RGB}{87,190,91}
\definecolor{brightgreen50}{RGB}{143,211,146}
\definecolor{brightgreen25}{RGB}{199,233,200}
\definecolor{brightgreen10}{RGB}{233,246,233}

\definecolor{raspberry100}{RGB}{130,0,78}
\definecolor{raspberry75}{RGB}{161,64,122}
\definecolor{raspberry50}{RGB}{193,128,167}
\definecolor{raspberry25}{RGB}{224,191,211}
\definecolor{raspberry10}{RGB}{243,230,237}

\definecolor{fuchsia100}{RGB}{198,0,126}
\definecolor{fuchsia75}{RGB}{212,64,158}
\definecolor{fuchsia50}{RGB}{227,128,191}
\definecolor{fuchsia25}{RGB}{241,191,223}
\definecolor{fuchsia10}{RGB}{249,230,242}

\definecolor{purple100}{RGB}{56,28,128}
\definecolor{purple75}{RGB}{106,85,160}
\definecolor{purple50}{RGB}{156,142,192}
\definecolor{purple25}{RGB}{205,198,223}
\definecolor{purple10}{RGB}{235,232,242}

\definecolor{amethyst100}{RGB}{117,59,189}
\definecolor{amethyst75}{RGB}{152,108,206}
\definecolor{amethyst50}{RGB}{186,157,222}
\definecolor{amethyst25}{RGB}{221,206,239}
\definecolor{amethyst10}{RGB}{241,235,248}

\definecolor{rust100}{RGB}{210,39,48}
\definecolor{rust75}{RGB}{221,93,100}
\definecolor{rust50}{RGB}{233,147,152}
\definecolor{rust25}{RGB}{244,201,203}
\definecolor{rust10}{RGB}{251,233,234}

\definecolor{tangerine100}{RGB}{237,112,0}
\definecolor{tangerine75}{RGB}{242,148,64}
\definecolor{tangerine50}{RGB}{246,184,128}
\definecolor{tangerine25}{RGB}{251,219,191}
\definecolor{tangerine10}{RGB}{253,241,230}

\definecolor{cobalt100}{RGB}{0,50,160}
\definecolor{cobalt75}{RGB}{64,101,184}
\definecolor{cobalt50}{RGB}{128,153,208}
\definecolor{cobalt25}{RGB}{191,204,231}
\definecolor{cobalt10}{RGB}{230,235,246}

\definecolor{azure100}{RGB}{0,113,222}
\definecolor{azure75}{RGB}{64,148,230}
\definecolor{azure50}{RGB}{128,184,239}
\definecolor{azure25}{RGB}{191,220,247}
\definecolor{azure10}{RGB}{230,241,252}

\usepackage{tabularx}
\usepackage{multicol}
\usepackage{placeins}
\usepackage{listings}
\usepackage{balance}
\newcolumntype{Y}{>{\raggedright\arraybackslash}X}

\lstdefinestyle{prompttemplate}{
    basicstyle=\fontsize{10}{12}\selectfont\ttfamily,
    columns=fullflexible,
    keepspaces=true,
    showstringspaces=false,
    upquote=true,
    breaklines=true,
    breakatwhitespace=true,
    breakautoindent=false,
    breakindent=0pt,
    frame=single,
    rulecolor=\color{black!35},
    framerule=0.4pt,
    framesep=8pt,
    xleftmargin=8.4pt,
    xrightmargin=8.4pt,
    aboveskip=8pt,
    belowskip=0pt,
    numbers=none,
    tabsize=2
}

\title{FIRME: Registry-Anchored Memory Graphs for Evolving, Heterogeneous Investor Knowledge}

\author{Yunbai Zhang}
\authornote{Equal contribution.}
\email{yunbai\_zhang@vanguard.com}
\affiliation{%
  \institution{The Vanguard}
  \city{Malvern}
  \state{Pennsylvania}
  \country{USA}}

\author{Yunyi Gao}
\authornotemark[1]
\email{yunyi\_gao@vanguard.com}
\affiliation{%
  \institution{The Vanguard}
  \city{Malvern}
  \state{Pennsylvania}
  \country{USA}}

\author{Han-Tai Shiao}
\authornote{Corresponding author.}
\email{han-tai\_shiao@vanguard.com}
\affiliation{%
  \institution{The Vanguard}
  \city{Malvern}
  \state{Pennsylvania}
  \country{USA}}
\renewcommand{\shortauthors}{Zhang, Gao, and Shiao}

\begin{document}
\begin{abstract}
    Financial AI workflows that reuse investor knowledge need persistent memory
    that integrates observations across sources and periods without losing identity
    or provenance. In a knowledge graph, recurring observations may use different
    surface forms, fragmenting relationship histories and preventing evidence from
    accumulating. We introduce Fusion and Identity Registry for Memory Evolution
    (FIRME), a registry-anchored memory graph that reconciles incoming observations
    to persistent canonical relationships and retains their source, time, and
    context. Within periods, rule-based fusion links investor-call discussions to
    portfolio cash-flow signals, directly or through a fixed product--topic
    ontology. Across periods, canonical identities and persistent aliases direct
    recurring observations to the same relationship history; deterministic updates
    summarize reinforcement, inactivity, and reactivation. We evaluate FIRME on
    proprietary investor-call and cash-flow records and on S\&P 500 earnings-call
    transcripts. Compared with a historical registry that retains canonical
    identities and semantic fallback but omits alias memory, FIRME increases
    final-period Correct Match Rate by 7.8--12.1 percentage points across four
    dataset--backend settings under a fixed-judge and fallback-review protocol.
    A blinded human audit of 200 incoming observations per setting finds Correct
    Match Rate gains of 9.0--16.5 points and reattachment-recall gains of
    22.5--29.0 points, with higher observed merge precision in all four samples.
    A separate audit finds valid evidence paths for 293 of 300 outputs from full
    within-period fusion in the investor-call setting. Extraction-faithfulness
    and round-trip diagnostics assess source grounding and preservation of
    extractor-derived graph content.
\end{abstract}

\maketitle

\begingroup
\renewcommand{\thefootnote}{}
\footnotetext{\copyright~2026 The Vanguard Group, Inc. All rights reserved.}
\footnotetext{The paper is accepted at the 3rd Workshop on LLMs and Generative AI for Finance (AI4F) at ACM ICAIF 2026 (non-archival).}
\endgroup

\section{Introduction}\label{sec:intro}

    Understanding an investor's interests requires attention to current evidence
    and to how those interests and relationships change with market conditions.
    The prominence of a concern such as market volatility may rise during
    turbulence, recede as conditions stabilize, and return when uncertainty
    increases. Reconstructing these trajectories therefore requires combining
    heterogeneous sources of evidence. Unstructured text, including call records
    and meeting notes, provides discussion context, whereas structured records,
    such as financial-activity tables, capture measurable behavior. Because
    interests can persist and return, past evidence should neither remain
    permanently active nor disappear when observations are absent.
    Agentic and tool-augmented financial workflows that reuse investor knowledge
    depend on maintaining this knowledge across interactions. This requires
    recurring observations to update the correct existing relationships while
    retaining their supporting evidence.

    Maintaining evidence across sources and periods requires reconciliation.
    Structured records and unstructured text differ in format and vocabulary;
    the same entity or relationship can vary in name or phrasing over time.
    Before entering the knowledge store, new evidence must therefore be aligned
    across sources and matched to the correct persistent identity. Otherwise,
    observations describing the same relationship form separate entities or
    connections, each with a partial history, while cross-format supporting
    evidence remains disconnected. Fragmented relationship histories in agent
    memory can yield incomplete or redundant context for downstream reasoning
    about an investor's recurring concerns. Stable identity is necessary to accumulate
    evidence under one persistent relationship and track its changes and
    lifecycle over time.

    Temporal graph memories provide an important starting point. Zep/Graphiti
    reconciles extracted entities with existing graph nodes, deduplicates facts,
    links facts to source episodes, and represents when facts are valid or
    invalidated; its evaluation emphasizes retrieval-supported question answering
    and latency~\cite{rasmussen2025zep}. In the financial settings we study, the
    central question is how evidence for the same relationship accumulates and
    changes across periods. Incoming observations may use different names or come
    from different sources, and some periods may contain no matched observation.
    The graph-update challenge is to determine which relationship history
    receives each new observation and how that relationship's evidence strength
    evolves with recurrence, inactivity, and renewed evidence.

    We introduce Fusion and Identity Registry for Memory Evolution (FIRME), a
    knowledge graph (KG) memory framework, to address this challenge. A stable
    head identifier anchors whose relationship is being tracked, while an evolving
    registry reconciles target-name variants to canonical identities, anchoring
    what that relationship concerns. Together, these identities determine which
    persistent relationship history receives incoming evidence. The memory graph
    retains each observation's relation label, source, time, and context. When
    conversational and structured evidence coexist, within-period fusion can
    recognize their corroboration while preserving both observations. Across
    periods, a recurring observation reinforces the corresponding history; periods
    without matched observations lower its evidence strength while leaving earlier
    evidence available. We test the incremental role of retained aliases against
    a control that still uses historical canonical reconciliation and semantic
    matching.

    FIRME is intended to provide investors and financial professionals with a
    consolidated, longitudinal view of investor-related information. By
    accumulating observations under persistent relationship identities, it
    brings current evidence together with earlier observations across sources,
    preserving their provenance and historical context.

    \subsection{Related Work}\label{sec:related}

        \textbf{Text-to-KG Construction and Canonicalization.}
        Text-to-KG construction progressed from pattern-based extraction and
        distant supervision~\cite{hearst1992automatic,mintz2009distant} to generative
        relation extraction~\cite{josifoski2022genie} and LLM pipelines. LLM pipelines
        add schema control and canonicalization: EDC separates open extraction,
        schema definition, and canonicalization, while KGGen clusters related
        entities~\cite{zhang2024edc,mo2025kggen}. GraphRAG structures extracted graphs
        for corpus-level summarization~\cite{edge2024local}, whereas open-KG methods
        consolidate synonymous entity and relation
        expressions~\cite{vashishth2018cesi,dash2021cuva}. These approaches reduce
        surface-form sparsity at construction time. They do not determine how
        recurring observations drawn from heterogeneous sources should accumulate
        under a single persistent relationship across successive time periods.

        \textbf{Temporal KGs and Dynamic Memory.}
        Temporal KGs model fact validity, while completion models predict missing or
        future facts from historical
        representations~\cite{leblay2018deriving,lacroix2020tensor,jin2020recurrent,li2021temporal}.
        ATOM separates observation time from validity time~\cite{lairgi2026atom}; Zep
        adds bi-temporal facts, provenance, resolution, and conflict-based
        invalidation for agent memory~\cite{rasmussen2025zep}; and FinDKG propagates
        latent representations across financial-news snapshots for
        prediction~\cite{li2024findkg}. These systems foreground fact validity,
        retrieval, or predictive state. We complement them by asking how recurring
        or absent evidence changes the salience and lifecycle of an explicit,
        auditable relationship edge retaining source, time, context, and
        participants. Recurrence strengthens its evidence state; non-observation
        lowers current prominence without erasing provenance; later evidence may
        reactivate it.

        \textbf{Heterogeneous Financial KG Fusion.}
        Multi-source KG integration commonly uses entity alignment and ontology
        matching~\cite{sun2020knowledge,otero2015ontology}. Financial KG research
        spans news extraction, temporal linkage, deduplication, conflict handling,
        table-aware construction, and evolving fundamentals or
        events~\cite{elhammadi2020financialkg,zhu2023hit,zhu2023tfinkb,arun2025finreflectkg,li2026finkario}.
        These systems primarily organize corporate news, disclosures, fundamentals,
        or discrete events. Investor-centric data require a different combination:
        conversational and structured observations may corroborate the same
        relationship without being duplicates, and missing evidence need not
        terminate that relationship, whose memory state may decay and later
        reactivate.

    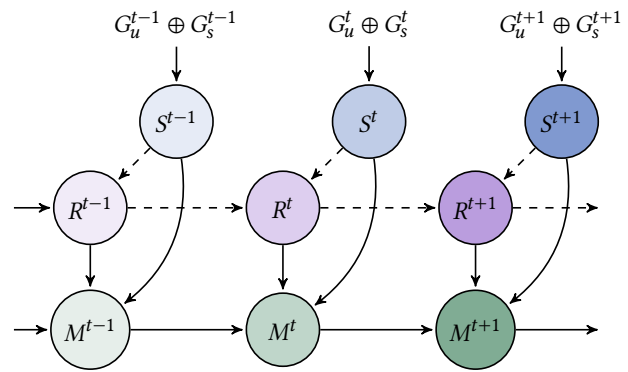
\begin{figure}[!t]
        \centering
        \resizebox{\linewidth}{!}{%
        \begin{tikzpicture}[->,>=stealth',shorten >=1pt,auto,node distance=2.5cm,semithick,initial text=New]
            \tikzstyle{every state}=[minimum size=3em]
            \node[state,fill=cobalt10]	(S1) {$S^{t-1}$};
            \node[state,fill=cobalt25]	(S2) [right of=S1] {$S^t$};
            \node[state,fill=cobalt50]	(S3) [right of=S2] {$S^{t+1}$};
            \node[] (G1) [above of=S1, node distance=4em] {$G_u^{t-1} \oplus G_s^{t-1}$};
            \node[] (G2) [above of=S2, node distance=4em] {$G_u^t \oplus G_s^t$};
            \node[] (G3) [above of=S3, node distance=4em] {$G_u^{t+1} \oplus G_s^{t+1}$};
            \node[state,fill=amethyst10] (R1) [below left of=S1, node distance=5em] {$R^{t-1}$};
            \node[state,fill=amethyst25] (R2) [below left of=S2, node distance=5em] {$R^t$};
            \node[state,fill=amethyst50] (R3) [below left of=S3, node distance=5em] {$R^{t+1}$};
            \node[state,fill=darkgreen10] (M1) [below of=R1, node distance=5em] {$M^{t-1}$};
            \node[state,fill=darkgreen25] (M2) [below of=R2, node distance=5em] {$M^{t}$};
            \node[state,fill=darkgreen50] (M3) [below of=R3, node distance=5em] {$M^{t+1}$};
            \node (leftR) [left of=R1, node distance=3.5em] {};
            \node (leftM) [left of=M1, node distance=3.5em] {};
            \node (rightR) [right of=R3, node distance=5.5em] {};
            \node (rightM) [right of=M3, node distance=5.5em] {};
            \path (G1) edge [] node {} (S1)
                  (G2) edge [] node {} (S2)
                  (G3) edge [] node {} (S3)
                  (S1) edge [dashed] node {} (R1)
                  (R1) edge [dashed] node {} (R2)
                  (S2) edge [dashed] node {} (R2)
                  (R2) edge [dashed] node {} (R3)
                  (R1) edge [] node {} (M1)
                  (M1) edge [] node {} (M2)
                  (S1) edge [bend left] node {} (M1)
                  (R2) edge [] node {} (M2)
                  (M2) edge [] node {} (M3)
                  (S2) edge [bend left] node {} (M2);
            \path (leftR) edge [] node {} (R1)
                  (leftM) edge [] node {} (M1)
                  (S3) edge [dashed] node {} (R3)
                  (R3) edge [dashed] node {} (rightR)
                  (R3) edge [] node {} (M3)
                  (M3) edge [] node {} (rightM)
                  (S3) edge [bend left] node {} (M3);
        \end{tikzpicture}
        }%
        \Description{In FIRME, across consecutive periods, the short-term graph and the previous registry jointly update the current registry (registry update); the short-term graph, the updated registry, and the previous memory graph are then reconciled to form the current memory graph (memory reconciliation).}
        \caption{Coupled short- and long-term fusion in FIRME across consecutive periods. Each short-term graph $S^t=G_u^t \oplus G_s^t$ fuses unstructured and structured graphs. Dashed arrows show registry update from $R^{t-1}$ to $R^t$, conditioned on $S^t$; solid arrows to $M^t$ show memory reconciliation of $S^t$, $R^t$, and $M^{t-1}$.}\label{fig:fusion}
    \end{figure}
    \FloatBarrier

    \subsection{Contributions}\label{sec:contributions}

    FIRME's central contribution is an identity-conditioned update of persistent
    relationship histories across periods (Figure~\ref{fig:fusion}). Within-period
    corroboration supports this update in paired investor-call data, while
    lifecycle states make the resulting histories interpretable.

    \textbf{Coupled registry--memory recurrence}.
    FIRME couples an evolving registry of canonical identities and retained
    aliases to a memory graph of relationship histories. A stable head identifier
    and a reconciled target identity determine which history receives each
    observation, even when names or evidence sources differ. The history retains
    each observation's relation label, source, timestamp, and context; it receives
    at most one strength and lifecycle update per period. Against a no-alias
    control that retains historical canonical reconciliation and semantic
    fallback, full FIRME improves final-period Correct Match Rate (CMR) by
    7.8--12.1 percentage points across four dataset--backend settings under the
    judge and fallback-review protocol (Table~\ref{tab:memory-usability}). In the
    shared blinded human audit (Table~\ref{tab:human-audit}),
    reattachment recall improves by 22.5--29.0 points, and observed merge
    precision is higher in all four samples.

    \textbf{Audited within-period corroboration}.
    For paired investor-call and cash-flow records, same-investor, same-period
    rules identify direct product corroboration and ontology-mediated support
    from a discussed topic. The derived product relationship references its
    supporting observations without merging topic and product identities. In
    the final-period Opus 4.6 human audit, 293 of 300 Full fusion outputs had
    valid evidence paths, including 91 of 94 outputs triggered by Rule~4 but
    not Rule~3 (Table~\ref{tab:fusion-audit}).

    \textbf{Interpretable lifecycle state assignment}.
    Given reconciled relationship histories, fixed rules update an
    evidence-strength score from matched or unmatched periods and assign one of
    five lifecycle states from match status, prior state, and consecutive
    unmatched periods. Each assigned state is traceable to retained observations
    and periods without a match, providing an interpretable summary of the
    relationship's recorded evidence trajectory.

    The longitudinal investor-call corpus includes paired cash-flow records and
    tests identity continuity and short-term corroboration. The transcript-only
    S\&P 500 corpus tests temporal reconciliation under a broader entity schema.
    Extraction-faithfulness and round-trip recovery diagnostics, lifecycle
    trajectories, and analytical parameter behavior provide complementary checks.

\section{Problem Formulation}\label{sec:problem}
    \subsection{Preliminaries and Notation}
        A knowledge graph is $G=(V,\mathcal{R},E)$, where $V$ is the entity
        set, $\mathcal{R}$ the relation-type set, and
        $E\subseteq V\times\mathcal{R}\times V$ contains directed, typed edges
        $(u,r,v)$ from head $u$ to tail $v$ through relation $r$. In the
        investor-call fusion setting, FIRME constructs a graph with the three
        entity types and three relation types defined in
        Section~\ref{sec:schema}. We call the product or topic at tail $v$ the
        \textit{target entity}; it is distinct from the head entity $u$. The
        S\&P 500 transcript-only setting uses a broader set of extracted
        entity types.

    \subsection{Data Modalities}
        In the investor-call setting, FIRME integrates two modalities. At period $t$,
        \textit{unstructured data} consists of text from call
        transcripts and meeting notes and yields $G_u^t$. Each tuple $(u,r,v)$
        relates a person or organization $u$ through $r$ to a product or topic
        $v$. The extractor is prompted to identify relationships involving
        substantive discussion.

        \textit{Structured data} consists of predefined records such as
        relational tables and yields $G_s^t$. A primary index identifies a
        person or organization; graph-construction rules specify which columns
        and value conditions instantiate edges. For example,
        an investor's capital inflow into a product can instantiate an
        investor--product behavioral edge.

    \subsection{Problem Statement}\label{sec:problem-statement}
        Constructing a temporally coherent graph from these inputs poses three
        interrelated problems.

        \textbf{Short-term fusion}.
        Given $G_u^t$ and $G_s^t$ within period $t$, short-term fusion combines
        the two partial graphs into the coherent, consolidated graph
        $S^t=G_u^t\oplus G_s^t$. Product- and topic-level evidence is aligned
        through a fixed product--topic ontology $O$
        (Appendix~\ref{app:ontology-examples}), while $\oplus$ combines
        discursive evidence with structured cash-flow signals. The
        \texttt{strong-interest} relation is assigned only when both modalities
        corroborate a relationship; it is a rule-based evidential label, not a
        calibrated preference or stance estimate.

        \textbf{Long-term fusion}.
        Long-term fusion maintains a temporally coherent memory graph as
        investment interests emerge, strengthen, persist, weaken, and may
        resurface after dormancy. Each canonical relationship in $M^t$ is a
        time-indexed edge whose attribute vector $\bm{h}_t$ accumulates
        observations $(u,r,v)$ and supporting evidence, together with first-
        and last-seen periods, a memory-strength score, and a lifecycle state.
        As Figure~\ref{fig:fusion} shows, $S^t$ is normalized through the
        prior registry $R^{t-1}$; these edges are
        reconciled with the prior memory graph $M^{t-1}$. These operations define
        \begin{equation}\label{eq:memory-update}
            \begin{aligned}
                R^t &= \Theta_1\!\left(R^{t-1}, S^t\right), \\
                M^t &= \Theta_2\!\left(M^{t-1}, R^t, S^t\right),
            \end{aligned}
        \end{equation}
        where $\Theta_1$ and $\Theta_2$ denote the coupled registry and memory
        reconciliation dynamics. We initialize $R^0=M^0=\varnothing$. At
        $t=1$, candidates in $S^1$ are deduplicated by lemmatized string
        equality and then by semantic matching for unresolved same-type
        candidates; all normalized edges enter $M^1$ in the \textit{Emerging}
        state with a baseline score of $0.5$. For $t>1$, unresolved same-type
        candidates are semantically compared only with $R^{t-1}$.

        \textbf{Entity normalization}.
        Entity normalization determines whether text-extracted entities $v_i$
        and $v_j$ in $G_u$ denote the same entity. Without it, near-duplicates
        fragment one relationship's history and burden both fusion stages;
        Section~\ref{sec:methodology} presents the registry-based solution.

\section{Methodology}\label{sec:methodology}
    \subsection{Entity and Relation Schema}\label{sec:schema}
        The investor-call fusion schema in Table~\ref{tab:schema} has three
        entity types: \texttt{person/org} for the person or organization whose
        relationships are tracked, \texttt{product} for a specific financial instrument, and
        \texttt{topic} for an investment theme or strategy. Three relation types
        link these head entities to products and topics. The \texttt{discussed} relation
        denotes a product or topic identified by the extractor as substantively
        discussed in the call. Structured
        \texttt{cf-signal} marks significant capital allocation toward a product,
        and \texttt{strong-interest} is a rule-derived evidential label requiring
        cross-modal corroboration, not a calibrated preference or stance estimate.

        The S\&P 500 transcript-only evaluation uses a broader extraction
        schema, with entity types such as Company, Person, Product,
        GeographicRegion, and FinancialMetric. These entities enter temporal
        reconciliation without the investor-call cash-flow fusion rules.

        The graph head is the tracked business entity: an investor or
        asset-management firm in the investor setting, or a reporting company
        in the S\&P 500 corpus. In the investor-call corpus, the call's canonical
        \texttt{subject\_entity\_id} and the cash-flow record's \texttt{owner}
        identify the same investor. The actual call speaker is recorded
        separately in \texttt{raw\_subject} and \texttt{speaker\_role}; a speaker
        change does not change the graph head. Cross-modal corroboration
        therefore requires the same investor ID and period, as well as the
        product or topic conditions in Section~\ref{sec:short-term-fusion}.

    \begin{table}[!t]
        \centering
        \small
        \setlength{\tabcolsep}{3pt}
        \renewcommand{\arraystretch}{0.96}
        \caption{Entity and relation schema for investor-call fusion.}
        \label{tab:schema}
\begin{tabularx}{\columnwidth}{@{}>{\raggedright\arraybackslash}p{0.36\columnwidth}Y@{}}
            \toprule
            \multicolumn{2}{@{}l}{\textbf{Entity types}} \\
            \midrule
            \texttt{person/org} & Tracked investor or asset-management firm \\
            \texttt{product}    & Financial instrument (e.g., \texttt{VOO}, \texttt{BND}) \\
            \texttt{topic}      & Investment theme (e.g., ESG Investing) \\
            \midrule
            \multicolumn{2}{@{}l}{\textbf{Relation types}} \\
            \midrule
            \shortstack[l]{\texttt{discussed}\\[-1pt]{\footnotesize\itshape Unstructured}}
                & Product/topic identified as substantively discussed \\
            \shortstack[l]{\texttt{cf-signal}\\[-1pt]{\footnotesize\itshape Structured}}
                & Investor shows significant inflow into a product \\
            \shortstack[l]{\texttt{strong-interest}\\[-1pt]{\footnotesize\itshape Fusion}}
                & Corroborating evidence in both modalities \\
            \bottomrule
        \end{tabularx}

    \end{table}

    \subsection{Knowledge Graph Construction}
        FIRME's text-to-graph component, \textit{LLM2Graph},
        builds the unstructured graph $G_u$ by processing each document
        $d\in D$ and emitting $(u,r,v,d)$, where $d$ records the supporting
        textual evidence. In the investor-call setting, it preserves each
        source-provided head ID and maps raw product and topic mentions to
        period-local normalized candidates,
        assigning each a deterministic identifier derived from its entity type
        and lemmatized name. The extractor's canonical layer performs
        period-local normalization without access to the historical entity registry
        or its learned aliases. The investor-call extraction system prompt is provided in
        Appendix~\ref{app:relationship-extraction-prompt}.

        After period-local graph construction and, when structured data are
        available, short-term fusion, the persistent registry maps candidates
        to cross-period canonical identities through deterministic-identifier,
        normalized-name, and alias matching. LLM-based semantic matching against
        canonical names and aliases is reserved for unresolved same-type
        candidates. Matched candidates adopt the anchor ID and add their surface
        forms to the anchor's alias set in $R^t$. Only candidates unresolved
        after all matching tiers retain their identifiers and extend the
        registry.

        The structured graph $G_s$ captures cash-flow signals, namely capital
        inflows from investors into products, as
        $(u,\texttt{cf-signal},v)$. Following the use-case definition of a
        material allocation, $G_s^t$ includes an edge only when monthly inflow
        exceeds \$100K; otherwise, the signal is ignored to avoid over-counting
        weak signals. Each record's \texttt{owner} field is the stable investor
        identifier.

    \subsection{Short-Term Fusion}\label{sec:short-term-fusion}
        Within each period, short-term fusion combines $G_u^t$ and $G_s^t$
        into $S^t$. Conversational product expressions are mapped to official
        product identities (e.g., \textit{Vanguard S\&P 500 ETF} to \textit{VOO}). We
        derived the product--topic ontology $O$ from our proprietary internal
        resource, manually annotated by our financial analyst team, and froze
        it before evaluation. The association
        semantics, representative records, and their use in Rule~4 are
        illustrated in Appendix~\ref{app:ontology-examples}.
        Original observations retain source indicator
        $1$ for unstructured evidence or $2$ for structured evidence.
        Corroborated product relationships receive the derived hybrid indicator
        $3$; this does not change the source indicators of their observations.

        Table~\ref{tab:fusion_rules} defines four rules. Rules~1--2 retain
        uncorroborated relationships with their original labels. Rule~3 performs
        direct corroboration: a current discussion and cash-flow signal for the
        same investor--product pair assign \texttt{strong-interest} to the
        product relationship. Rule~4 combines a current topic discussion with a
        current cash-flow signal for an ontology-linked product from the same
        investor. The product relationship is labeled \texttt{strong-interest},
        while the topic relationship remains \texttt{discussed}. A topic may
        corroborate several allocated products without being duplicated in its
        own history.

        Fusion retains the original targets and provenance of all observations.
        Under Rule~3, both observations belong to the same product history;
        under Rule~4, the product relationship references the supporting topic
        discussion, which also remains in its own topic history. All applicable
        direct and ontology-mediated support paths are retained as references,
        not additional observations. All original targets remain available for
        registry reconciliation. Appendix~\ref{app:fusion-memory-examples}
        provides pseudocode and cross-period examples.

    \begin{table*}[htbp]
        \centering
        \small
        \setlength{\tabcolsep}{3pt}
        \caption{Short-term fusion rules in FIRME. Rules~1--2 retain uncorroborated
        relationships; Rules~3--4 assign \texttt{strong-interest} to products
        using current-period evidence from the same investor. Rule~4 retains
        the discussed-topic relationship. Original observations and provenance
        are preserved under all rules; \texttt{TaxEff} denotes Tax Efficiency.}
        \label{tab:fusion_rules}
\begin{tabularx}{\textwidth}{@{}ccYY@{}}
            \toprule
            \textbf{Rule} & \textbf{Type} & \textbf{Antecedent} & \textbf{Consequent} \\
            \midrule
            1 & \texttt{Pass} & \texttt{discussed(Alice, VFIAX, 1)} & $\Rightarrow$ \texttt{discussed(Alice, VFIAX, 1)} \\
            2 & \texttt{Pass} & \texttt{cf-signal(Bob, VOO, 2)}      & $\Rightarrow$ \texttt{cf-signal(Bob, VOO, 2)} \\
            3 & \texttt{Fuse} & \texttt{discussed(Chris, BND, 1)} $\oplus$\newline \texttt{cf-signal(Chris, BND, 2)}
              & $\Rightarrow$ \texttt{strong-interest(Chris, BND, 3)} \\
            4 & \texttt{Fuse} & \texttt{discussed(Emma, TaxEff, 1)} $\oplus$\newline \texttt{cf-signal(Emma, VTEB, 2)},\newline with $(\texttt{VTEB},\texttt{TaxEff})\in O$
              & $\Rightarrow$ \texttt{discussed(Emma, TaxEff, 1)}\newline and \texttt{strong-interest(Emma, VTEB, 3)} \\
            \bottomrule
        \end{tabularx}

    \end{table*}

    \subsection{Long-Term Memory Fusion}\label{sec:long-term-fusion}
        Long-term memory fusion combines registry-based entity reconciliation
        with rule-based edge reconciliation. Entity reconciliation establishes
        persistent identities, while edge reconciliation accumulates observations
        and updates relationship histories, memory strength, and lifecycle states
        across periods.

        \textbf{Cross-Period Entity Reconciliation}.
        FIRME resolves incoming entities against the prior registry through
        identifier, normalized-name, and learned-alias matching, applied as
        sequential deterministic checks. That is, it first checks for an exact
        identifier match, then compares the incoming entity's normalized name
        with registered canonical names, and finally checks for a matching alias
        previously associated with a canonical identity. The sequence stops once
        a match is found. For each unresolved entity, the candidate set comprises
        all entities of the same type in the prior registry $R^{t-1}$.
        The LLM assigns semantic similarity scores to these candidates using
        their canonical names and aliases. The system prompt for semantic
        matching is provided in Appendix~\ref{app:semantic-entity-matching-prompt}.
        A deterministic rule filters candidates with scores
        $\geq 0.8$ and selects the highest-scoring candidate. The incoming entity
        adopts the selected canonical ID, and its surface forms are added to that
        identity's alias set. If no candidate meets the threshold, the entity is
        registered as new. An exploratory threshold-selection analysis is
        provided in Appendix~\ref{app:semantic-threshold}.
        Appendix~\ref{app:entity-reconciliation} provides pseudocode.

        \textbf{Cross-Period Edge Reconciliation}.
        After entity reconciliation, each original observation updates the
        history identified by its head ID and typed canonical target ID.
        Topic and product identities remain distinct; relation labels and
        source indicators are preserved per observation and excluded from the
        key. Each key undergoes one update per period: initialization if new,
        reinforcement if observed again, or decay if no current observation
        targets it. Multiple observations or corroboration paths do not produce
        additional score updates. Historical corroboration remains retrievable
        but does not count as current evidence. Relation upgrades follow
        Table~\ref{tab:fusion_rules} independently of memory-strength updates.

        Within this procedure, the \textit{Memory Reinforcement--Decay Model}
        updates the memory-strength score $\sigma_t(e)\in[0,1]$ through a
        deterministic match gate. It draws on the gating intuition of long
        short-term memory (LSTM) networks~\cite{hochreiter1997long}, rather than
        adopting their learned recurrent architecture.
        A new edge starts at $\sigma_{\mathrm{init}}=0.5$. An existing matched
        edge is reinforced additively at rate $\alpha=0.2$,
        \begin{equation}\label{eq:sigma_update}
            \sigma_t(e) = \min\!\left(1.0,\; \sigma_{t-1}(e) + \alpha\right),
        \end{equation}
        driving repeated matches toward saturation. An unmatched
        edge decays multiplicatively at retention rate $\gamma=0.8$,
        \begin{equation}\label{eq:sigma_decay}
            \sigma_t(e) = \gamma\,\sigma_{t-1}(e),
        \end{equation}
        giving exponential decay. Scores below $0.01$ become zero, but the edge
        remains stored.

        \textbf{Lifecycle States}.
        Lifecycle states provide an interpretable summary of each relationship's
        temporal trajectory, determined by match status, prior state, and an inactivity counter for
        consecutive unmatched periods. An edge is \textit{Emerging} when newly
        created; \textit{Reactivated} when matched after
        \textit{Fading} or with prior inactivity above $N$;
        \textit{Reinforced} for other matches;
        \textit{Stable} when unmatched with inactivity at most $N$; and
        \textit{Fading} when unmatched with inactivity above $N$.

\begin{figure}[!t]
        \centering
        \resizebox{0.78\columnwidth}{!}{%
        \begin{tikzpicture}[->,>=stealth',shorten >=1pt,auto,
            node distance=2.5cm,semithick,initial text=New,
            every state/.style={draw,circle,fill=orange!25,minimum size=1.15cm}]
            \node[state,initial] (E) {Emerg.};
            \node[state] (S) [above right of=E] {Stable};
            \node[state] (R) [below right of=E] {Reinf.};
            \node[state] (F) [right of=S] {Fading};
            \node[state] (A) [right of=R] {React.};
            \path (E) edge [bend left,dashed] node {} (S)
                  (S) edge [bend left,dashed]
                      node[above,font=\footnotesize] {$>3$ misses} (F)
                  (S) edge [bend right=20] node {} (R)
                  (E) edge [bend right] node {} (R)
                  (R) edge [bend right=20,dashed] node {} (S)
                  (A) edge [bend left] node {} (R)
                  (A) edge [dashed] node {} (S)
                  (F) edge [bend left=20] node {} (A)
                  (R) edge [loop below] node {} (R)
                  (S) edge [loop above,dashed]
                      node[above,font=\footnotesize] {$\leq 3$ misses} (S)
                  (F) edge [loop above,dashed] node {} (F);
        \end{tikzpicture}%
        }
        \Description{A five-state lifecycle diagram. New relationships enter
        Emerging. A match leads to Reinforced unless the prior state is Fading,
        in which case it leads to Reactivated. Consecutive misses lead to Stable
        through the third miss and to Fading on the fourth. Initialization and
        matches reset the inactivity count. Solid transitions denote matches;
        dashed transitions denote misses.}
        \caption{Memory-edge lifecycle with inactivity threshold $N=3$.
        The entry arrow initializes a relationship as Emerging. Solid
        transitions between states, including self-loops, denote matches;
        dashed transitions denote misses. Guards count consecutive misses
        including the current period; a match resets the count.}
        \label{fig:state-machine}
    \end{figure}
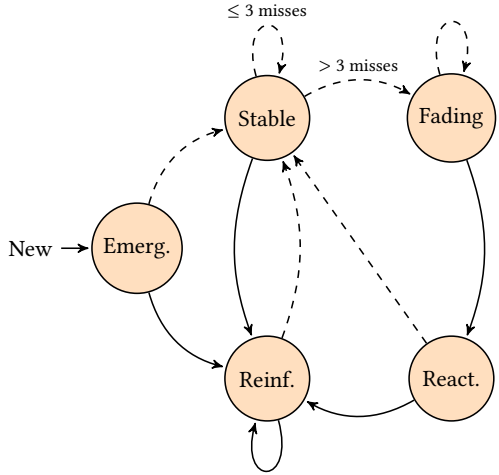

        Initialization and matches reset the inactivity counter to zero;
        misses increment it. With $N=3$, an edge remains \textit{Stable} through
        three consecutive unmatched periods and becomes \textit{Fading} on the
        fourth (Figure~\ref{fig:state-machine}). Scores decay from the first
        unmatched period, including periods in the \textit{Stable} state.
        Appendix~\ref{app:fusion-memory-examples} provides worked examples.
        The resulting
        state estimates interest evolution from observed evidence rather than
        measuring latent preference directly.

        Entity reconciliation and edge reconciliation implement $\Theta_1$ and
        $\Theta_2$, respectively, in Equation~\ref{eq:memory-update}. Edges are
        never removed; even zero-score edges remain available for reactivation.

        \textbf{Relationship as an Attribute Vector}.
        Whereas the base relation $r$ in a static triple $(u,r,v)$ is a single
        categorical label, its counterpart in $M^t$ is a time-dependent
        attribute vector,
        \begin{equation}
            e_t = \left(u,\, \bm{h}_t,\, v\right),
        \end{equation}
        where $\bm{h}_t$ stores each observation's identifier, original typed
        target, relation type $r$, source indicator, timestamp, supporting
        context, and head identifier, plus speaker identity and role when
        available, together with edge-level
        first- and last-seen periods, memory state, and score $\sigma_t(e)$.
        Corroboration records reference supporting observations and applicable
        ontology links without changing the observations' original targets.
        Observations reconciled to the same
        canonical edge accumulate in $\bm{h}_t$ without overwriting earlier
        evidence, so score and state summarize the relationship's evolving
        strength and lifecycle.

\section{Empirical Evaluation}\label{sec:eval}
    \subsection{Evaluation Criteria and Setup}
        We evaluate FIRME primarily through \textbf{memory usability}, focusing
        on cross-period identity continuity and graph consolidation.
        \textbf{Extraction faithfulness} and \textbf{round-trip recovery}
        provide supporting diagnostics of source support and extractor-derived
        content preservation after graph-to-text-to-graph conversion.
        A final human audit evaluates the evidence supporting direct and
        ontology-mediated product corroboration.

        We evaluate on two corpora. The first is a proprietary corpus of real investor-call transcripts paired with monthly portfolio cash-flow records; it spans 10 monthly periods with roughly 1,200 calls each. The transcripts and cash-flow records were obtained from routine business records and linked by investor identifier and month. Any personally identifiable information (PII) was masked. These data are not publicly available due to confidentiality restrictions.

        The second corpus is the S\&P 500 Earnings Transcripts
        Dataset~\cite{bose345sp500earnings}. We sampled 297 reporting
        companies with earnings calls dated February 2025 in the final period
        $T$ and used their available transcripts across three reporting periods.
        Extracted entity types include companies, people, products and
        services, geographic regions, and financial topics.
        Each reporting company ID serves as a graph head. Because this
        corpus contains only transcripts, it evaluates extraction and
        temporal reconciliation, not cash-flow fusion.

        Memory usability, extraction faithfulness, and round-trip recovery
        use evaluation inputs from the final period $T$. The final period
        contains approximately 1,300 investor calls with 10.3K incoming edges,
        and 297 S\&P 500 earnings calls with approximately 6.3K incoming edges.
        The approximate edge totals refer to the Opus 4.6 reconciliation
        inputs. Round-trip recovery samples reference graphs from these
        final-period extracted outputs.

        Unless stated otherwise, primary language-model components use Claude
        Opus 4.6 (Opus 4.6), and both corpora are additionally evaluated with
        GPT-5.4-mini; both backends use temperature 0.0. The
        extraction-faithfulness benchmark includes
        GraphRAG~\cite{edge2024local} as an additional comparison framework.
        Gemini 3.1 Pro serves as the fixed LLM judge across both
        datasets and all evaluated backends and strategies. It assesses
        grounding, normalization accuracy, and
        adjusted-precision grounding in round-trip recovery.
        Reattachment correctness is assessed through two complementary
        protocols. For LLM-Norm, FIRME w/o Alias Memory, and FIRME,
        Gemini 3.1 Pro evaluates all accepted reattachments, with two annotators
        reviewing cases initially judged incorrect
        (Appendix~\ref{app:evaluation-details}). A separate blinded human
        audit establishes an adjudicated reference for 200 incoming units per
        dataset--backend setting, shared by all five strategies
        (Appendix~\ref{app:human-audit}).
        Memory updates use reinforcement
        $\alpha=0.2$, per-period retention $\gamma=0.8$, initialization
        $\sigma_{\mathrm{init}}=0.5$, inactivity threshold $N=3$, and zeroing
        threshold 0.01. Deterministic identifier and name matches are accepted
        directly; unresolved candidates enter a semantic tier at threshold 0.8.
        This threshold was selected using the penultimate period ($T-1$) of the
        investor-call corpus (Appendix~\ref{app:semantic-threshold}) and used for
        final-period evaluation.

    \subsection{Memory Usability}
        We first evaluate whether registry-anchored reconciliation enables
        recurring observations to accumulate under persistent relationship
        identities. We define
        \textit{identity continuity} as linking semantically equivalent
        observations across periods to the same canonical relationship. We
        compare \textbf{RAW} exact surface forms, \textbf{spaCy-Lemma}
        deterministic lemmatization~\cite{honnibal2020spacy},
        \textbf{LLM-Norm} independent within-period semantic normalization,
        \textbf{FIRME w/o Alias Memory}, and full \textbf{FIRME}.
        LLM-Norm normalizes each period independently, after which equal
        names or identifiers can be matched deterministically across periods.
        To evaluate the incremental contribution of persistent alias memory,
        we introduce FIRME w/o Alias Memory as a canonical-only historical
        registry control. It retains canonical-ID and canonical-name matching
        and cross-period semantic fallback, but removes the storage and reuse
        of alternative names. Appendix~\ref{app:alias-memory-control} details
        the control and the role of each comparison.

        The \textbf{Edge Compression Rate} and \textbf{Entity Compression Rate}
        measure reductions in distinct edges and entities in cumulative memory
        relative to RAW; higher values indicate greater consolidation. The
        \textbf{Memory Match Rate (MMR)} is the fraction of incoming observations
        represented in $S^t$ that are reattached to cumulative memory $M^{t-1}$, and
        \textbf{Merge Precision} measures the proportion of accepted
        reattachments judged to target the correct existing relationship.
        \textbf{Correct Match Rate (CMR)} is the proportion of incoming observations
        judged to be correctly reattached to cumulative prior memory:
        \begin{equation}\label{eq:correct-match-rate}
            \mathrm{CMR} = \mathrm{MMR} \times \text{Merge Precision}.
        \end{equation}
        Within each dataset--backend setting, strategies share the same
        incoming observations. We evaluate whether each incoming observation
        is correctly reattached. Multiple observations can refer to the same
        historical relationship, so they remain separate evaluation units
        while contributing to one relationship history. CMR uses all incoming
        observations as its denominator. Human-audit reattachment recall instead measures the
        fraction of gold-recurring incoming units correctly reattached to
        their historical identities.

        Compression rates describe cumulative memory at the final period;
        MMR covers all final-period
        incoming observations. In Table~\ref{tab:memory-usability}, precision and CMR
        for LLM-Norm, FIRME w/o Alias Memory, and FIRME use the judge and
        fallback-review protocol. RAW and spaCy-Lemma use backend-specific
        human-audit precision; their CMR estimates multiply the reported
        population MMR by that precision. The final column reports
        human-audit reattachment recall. Appendix~\ref{app:human-audit}
        provides the complete human-audit results.

\begin{figure}[!t]
    \centering
    \begin{tikzpicture}[
        every node/.style = {font=\small},
        hub/.style       = {circle, thick, draw=cobalt100, fill=cobalt10, minimum size=9mm, inner sep=1pt},
        leaf/.style      = {circle, thick, minimum size=5mm, inner sep=0pt},
        link/.style      = {thick, black},
        caption/.style   = {font=\small, anchor=north},
    ]

    \begin{scope}
        \node[hub] (cT) at (0,0) {investor};
        \node[leaf, draw=black,     fill=white]       (T1)  at (  0.0 : 1.4cm) {};
        \node[leaf, draw=tangerine100, fill=tangerine25] (T2)  at ( 32.7 : 2.0cm) {};
        \node[leaf, draw=darkgreen100, fill=darkgreen25] (T3)  at ( 65.5 : 1.5cm) {};
        \node[leaf, draw=tangerine100, fill=tangerine25] (T4)  at ( 98.2 : 2.0cm) {};
        \node[leaf, draw=amethyst100,  fill=amethyst25]  (T5)  at (130.9 : 1.7cm) {};
        \node[leaf, draw=darkgreen100, fill=darkgreen25] (T6)  at (163.6 : 1.5cm) {};
        \node[leaf, draw=amethyst100,  fill=amethyst25]  (T7)  at (196.4 : 1.7cm) {};
        \node[leaf, draw=amethyst100,  fill=amethyst25]  (T8)  at (229.1 : 1.7cm) {};
        \node[leaf, draw=darkgreen100, fill=darkgreen25] (T9)  at (261.8 : 2.0cm) {};
        \node[leaf, draw=black,     fill=white]       (T10) at (294.5 : 1.4cm) {};
        \node[leaf, draw=tangerine100, fill=tangerine25] (T11) at (327.3 : 2.0cm) {};

        \node [align=left]   at (T1.east)  [anchor=west]  {active\\management};
        \node [align=left]   at (T2.east)  [anchor=west]  {municipal\\bond};
        \node                at (T3.north) [anchor=south] {tlh};
        \node                at (T4.north) [anchor=south] {muni bond};
        \node [align=right]  at (T5.west)  [anchor=east]  {extending bond\\duration};
        \node [align=right]  at (T6.west)  [anchor=east]  {tax-loss\\harvesting};
        \node [align=right]  at (T7.west)  [anchor=east]  {lengthen\\duration};
        \node [align=right]  at (T8.west)  [anchor=east]  {extend\\duration};
        \node [align=center] at (T9.south) [anchor=north] {harvesting\\tax losses};
        \node [align=left]   at (T10.south east) [anchor=north west] {investment-grade\\treasury};
        \node [align=left]   at (T11.east) [anchor=west]  {munis};

        \foreach \n in {1,...,11} { \draw[link] (cT) -- (T\n); }
    \end{scope}

    \begin{scope}[yshift=-6.4cm]
        \node[hub] (cB) at (0,0) {investor};

        \node[leaf, draw=black,     fill=white]       (B1) at (  0.0 : 1.4cm) {};
        \node[leaf, draw=black,     fill=white]       (B2) at (294.5 : 1.4cm) {};
        \node[leaf, draw=tangerine100, fill=tangerine25] (B3) at ( 98.2 : 2.0cm) {};
        \node[leaf, draw=darkgreen100, fill=darkgreen25] (B4) at (163.6 : 1.5cm) {};
        \node[leaf, draw=amethyst100,  fill=amethyst25]  (B5) at (196.4 : 1.7cm) {};

        \node [align=left]  at (B1.east) [anchor=west] {active\\management};
        \node [align=left]  at (B2.east) [anchor=west] {investment-grade\\treasury};
        \node [align=right] at (B3.west) [anchor=east] {municipal\\bond};
        \node [align=right] at (B4.west) [anchor=east] {tax-loss\\harvesting};
        \node [align=right] at (B5.west) [anchor=east] {duration\\extension};

        \foreach \n in {1,...,5} { \draw[link] (cB) -- (B\n); }
    \end{scope}

    \node[caption] at (0,-3.3) {(a) LLM-Norm};
    \node[caption] at (0,-8.4) {(b) FIRME};
    \end{tikzpicture}

    \Description{Two investor-centered topic graphs compare 11 surface-form topics under LLM-Norm with five canonical topics under FIRME. Colors indicate which surface variants are consolidated, while uncolored topics remain distinct.}
    \caption{Topic consolidation under LLM-Norm~(a) and
    FIRME~(b); colors link surface variants assigned to one
    canonical topic, while uncolored topics remain distinct.}
    \label{fig:topic-consolidation}
\end{figure}
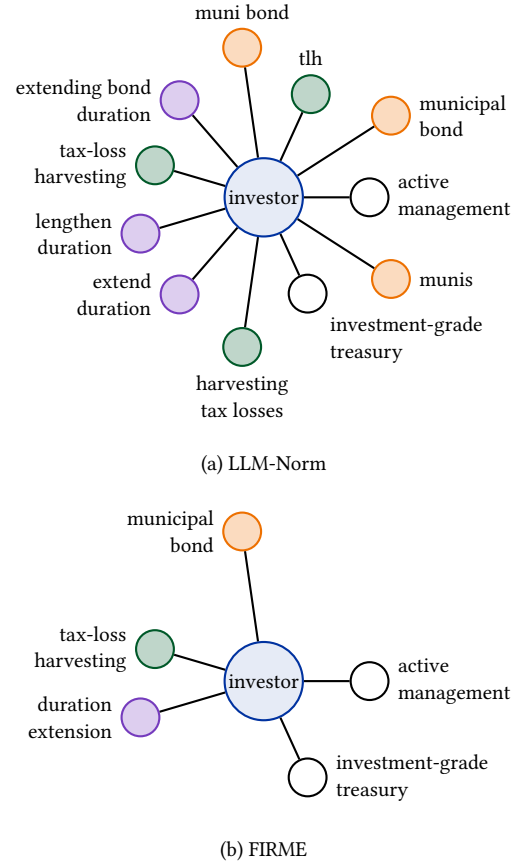

    \begin{table*}[!t]
        \centering
        \small
        \setlength{\tabcolsep}{2.5pt}
        \renewcommand{\arraystretch}{1.12}
        \caption{Memory usability. CR (compression rates) and MMR summarize
        population outputs. Precision uses human-audit estimates for
        RAW/spaCy-Lemma ($\dagger$); otherwise, Gemini 3.1 Pro judges all accepted
        reattachments, with human review of initially incorrect cases.
        CMR is population MMR $\times$ precision. All reattachment recall values
        come from the shared human audit (200 incoming units per setting;
        CIs in Appendix~\ref{app:human-audit}). Bold marks the highest CR, MMR,
        CMR, and recall within each setting.}
        \label{tab:memory-usability}
        \begin{tabularx}{\textwidth}{@{}lll*{5}{>{\centering\arraybackslash}X}>{\centering\arraybackslash}p{0.13\textwidth}@{}}
        \toprule
        \textbf{Dataset} & \textbf{Backend} & \textbf{Strategy} &
        \textbf{\shortstack{Edge\\CR}} & \textbf{\shortstack{Entity\\CR}} &
        \textbf{MMR} & \textbf{\shortstack{Merge\\precision}} & \textbf{CMR} &
        \textbf{\shortstack{Reattachment\\Recall}} \\
        \cmidrule(lr){4-8}\cmidrule(l){9-9}
        \multirow{5}{*}{\shortstack{Investor\\calls}} & \multirow{5}{*}{Opus 4.6} & RAW & 0.0\% & 0.0\% & 7.6\% & 100.0\%$^\dagger$ & 7.6\%$^\dagger$ & 15.0\% \\
         &  & spaCy-Lemma & 3.0\% & 7.6\% & 8.4\% & 94.1\%$^\dagger$ & 7.9\%$^\dagger$ & 20.0\% \\
         &  & LLM-Norm & 6.1\% & 22.6\% & 8.3\% & 98.5\% & 8.2\% & 20.0\% \\
         &  & \shortstack[l]{FIRME w/o\\Alias Memory} & 15.4\% & 43.7\% & 24.1\% & 92.3\% & 22.2\% & 52.5\% \\
         &  & \textbf{Full FIRME} & \textbf{20.2\%} & \textbf{51.1\%} & \textbf{31.8\%} & 95.5\% & \textbf{30.4\%} & \textbf{77.5\%} \\
        \midrule
        \multirow{5}{*}{\shortstack{Investor\\calls}} & \multirow{5}{*}{\shortstack{GPT-5.4-\\mini}} & RAW & 0.0\% & 0.0\% & 5.9\% & 100.0\%$^\dagger$ & 5.9\%$^\dagger$ & 15.0\% \\
         &  & spaCy-Lemma & 1.9\% & 4.2\% & 6.6\% & 100.0\%$^\dagger$ & 6.6\%$^\dagger$ & 16.3\% \\
         &  & LLM-Norm & 10.0\% & 26.5\% & 9.0\% & 83.5\% & 7.5\% & 20.0\% \\
         &  & \shortstack[l]{FIRME w/o\\Alias Memory} & 16.2\% & 35.2\% & 22.4\% & 83.3\% & 18.7\% & 46.3\% \\
         &  & \textbf{Full FIRME} & \textbf{20.7\%} & \textbf{41.4\%} & \textbf{31.9\%} & 83.0\% & \textbf{26.5\%} & \textbf{68.8\%} \\
        \midrule
        \multirow{5}{*}{S\&P 500} & \multirow{5}{*}{Opus 4.6} & RAW & 0.0\% & 0.0\% & 19.0\% & 100.0\%$^\dagger$ & 19.0\%$^\dagger$ & 21.5\% \\
         &  & spaCy-Lemma & 4.7\% & 6.9\% & 25.3\% & 98.0\%$^\dagger$ & 24.8\%$^\dagger$ & 38.5\% \\
         &  & LLM-Norm & 3.9\% & 6.0\% & 24.1\% & 99.3\% & 23.9\% & 36.2\% \\
         &  & \shortstack[l]{FIRME w/o\\Alias Memory} & 8.8\% & 14.7\% & 49.2\% & 93.8\% & 46.1\% & 63.1\% \\
         &  & \textbf{Full FIRME} & \textbf{10.5\%} & \textbf{17.9\%} & \textbf{59.1\%} & 97.2\% & \textbf{57.4\%} & \textbf{88.5\%} \\
        \midrule
        \multirow{5}{*}{S\&P 500} & \multirow{5}{*}{\shortstack{GPT-5.4-\\mini}} & RAW & 0.0\% & 0.0\% & 14.1\% & 100.0\%$^\dagger$ & 14.1\%$^\dagger$ & 21.0\% \\
         &  & spaCy-Lemma & 2.3\% & 5.2\% & 16.3\% & 97.0\%$^\dagger$ & 15.8\%$^\dagger$ & 32.0\% \\
         &  & LLM-Norm & 2.8\% & 7.1\% & 16.2\% & 99.5\% & 16.1\% & 30.0\% \\
         &  & \shortstack[l]{FIRME w/o\\Alias Memory} & 5.4\% & 14.7\% & 28.6\% & 86.0\% & 24.6\% & 48.0\% \\
         &  & \textbf{Full FIRME} & \textbf{7.6\%} & \textbf{18.2\%} & \textbf{44.8\%} & 82.0\% & \textbf{36.7\%} & \textbf{77.0\%} \\
        \bottomrule
        \end{tabularx}
    \end{table*}

    \newcommand{\lifecyclefigure}{%
    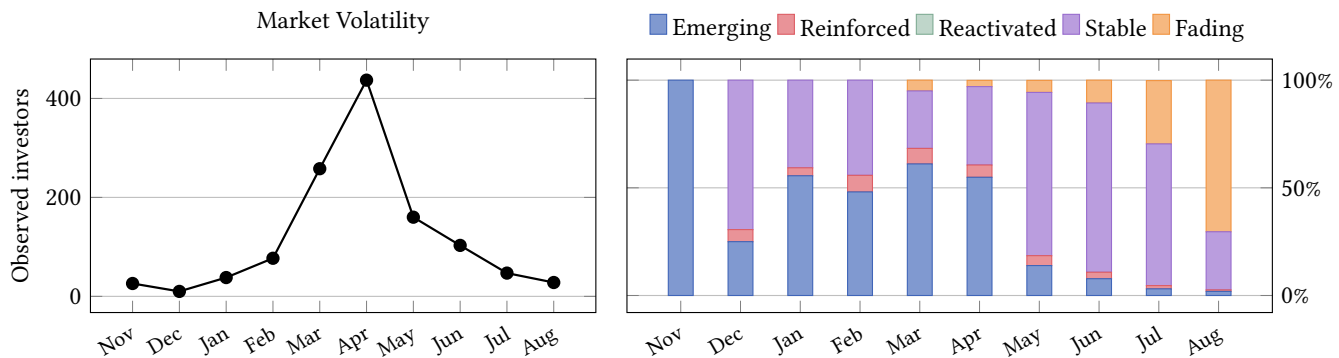
\begin{figure*}[!t]
            \centering
\resizebox{\textwidth}{!}{%
\begin{tikzpicture}
    \begin{axis}[
        name=lineaxis,
        width=0.45\textwidth,
        height=0.27\textwidth,
        symbolic x coords={Nov,Dec,Jan,Feb,Mar,Apr,May,Jun,Jul,Aug},
        xtick=data,
        xticklabel style={font=\small, rotate=30, anchor=north east},
        ylabel=Observed investors,
        ymajorgrids,
        title={Market Volatility},
        title style={at={(0.5,1.0)}, anchor=south},
        ]
        \addplot [color=black,mark=*,thick] coordinates {
            (Nov, 26)
            (Dec, 10)
            (Jan, 38)
            (Feb, 77)
            (Mar, 258)
            (Apr, 437)
            (May, 160)
            (Jun, 103)
            (Jul, 47)
            (Aug, 28)
            };
    \end{axis}
    \begin{axis}[
        at={($(lineaxis.south east)+(1.25em,0)$)},
        anchor=south west,
        ybar stacked,
        width=0.55\textwidth,
        height=0.27\textwidth,
        symbolic x coords={Nov,Dec,Jan,Feb,Mar,Apr,May,Jun,Jul,Aug},
        xtick=data,
        xticklabel style={font=\small, rotate=30, anchor=north east},
        yticklabel pos=right,
        yticklabel={\pgfmathprintnumber{\tick}\%},
        ylabel near ticks,
        ymajorgrids,
        bar width=1em,
        legend style={at={(0.5,1.03)}, anchor=south, legend columns=-1, draw=none},
        ]
        \addplot+[ybar] [color=cobalt75,fill=cobalt50] coordinates {
            (Nov, 100)
            (Dec, 25)
            (Jan, 55.6)
            (Feb, 48.1)
            (Mar, 61.1)
            (Apr, 54.9)
            (May, 13.9)
            (Jun, 7.8)
            (Jul, 3.1)
            (Aug, 1.9)
            };
        \addplot+[ybar] [color=rust75,fill=rust50] coordinates {
            (Nov, 0)
            (Dec, 5.6)
            (Jan, 3.7)
            (Feb, 7.7)
            (Mar, 7.2)
            (Apr, 5.7)
            (May, 4.6)
            (Jun, 3.0)
            (Jul, 1.5)
            (Aug, 0.7)
            };
        \addplot+[ybar] [color=darkgreen50,fill=darkgreen25] coordinates {
            (Nov, 0)
            (Dec, 0)
            (Jan, 0)
            (Feb, 0)
            (Mar, 0)
            (Apr, 0)
            (May, 0)
            (Jun, 0)
            (Jul, 0)
            (Aug, 0)
            };
        \addplot+[ybar] [color=amethyst75,fill=amethyst50] coordinates {
            (Nov, 0)
            (Dec, 69.4)
            (Jan, 40.7)
            (Feb, 44.2)
            (Mar, 26.7)
            (Apr, 36.4)
            (May, 75.8)
            (Jun, 78.6)
            (Jul, 65.8)
            (Aug, 27)
            };
        \addplot+[ybar] [color=tangerine75,fill=tangerine50] coordinates {
            (Nov, 0)
            (Dec, 0)
            (Jan, 0)
            (Feb, 0)
            (Mar, 5.0)
            (Apr, 2.9)
            (May, 5.6)
            (Jun, 10.6)
            (Jul, 29.4)
            (Aug, 70.4)
            };
        \legend{Emerging, Reinforced, Reactivated, Stable, Fading}
    \end{axis}
\end{tikzpicture}
}%
            \Description{Two-panel figure for the Market Volatility topic group on the investor-call corpus from November 2024 to August 2025. Panel (a) is a line chart of monthly observed active investors; panel (b) is a stacked bar chart showing the shares of lifecycle states (Emerging, Reinforced, Reactivated, Stable, Fading) over accumulated memory edges.}
            \caption{Market Volatility topic group in the investor-call corpus,
            November 2024--August 2025: monthly observed investors (left) and
            lifecycle-state shares over accumulated memory edges (right).}
            \label{fig:pattern}
        \end{figure*}%
    }

        Under the judge and fallback-review protocol, full FIRME increases
        CMR over FIRME w/o Alias Memory by 7.8--12.1 percentage points across
        the four settings (Table~\ref{tab:memory-usability}). These gains
        support the incremental contribution of persistent alias memory
        beyond canonical-only historical reconciliation. Full FIRME also
        achieves 2.3--3.7 times the CMR of LLM-Norm. The canonical-only control
        exceeds LLM-Norm in every setting, supporting the combined value of
        historical canonical reconciliation and semantic fallback.

        In this evaluation, merge precision improves with alias memory in
        both Opus 4.6 settings and changes little on investor calls with
        GPT-5.4-mini. On S\&P 500 with GPT-5.4-mini, CMR increases from
        24.6\% to 36.7\%, while precision decreases from 86.0\% to 82.0\%.
        These judge-based results indicate a coverage--precision trade-off
        in that setting.

        FIRME also yields the highest Edge Compression Rate and Entity
        Compression Rate in every setting, providing complementary evidence
        of reduced memory fragmentation.
        Figure~\ref{fig:topic-consolidation} illustrates the reduction from
        11 surface-form topics under LLM-Norm to five canonical topics under
        FIRME, while preserving distinct topics.

        \textbf{Human-audit methods.} Because population coverage statistics
        do not establish correctness or missed matches, we complemented them
        with a blinded human audit. We randomly sampled 200 final-period
        incoming relationship units per dataset--backend setting without
        replacement, including unmatched units, and evaluated all five
        strategies against the same reference. Two annotators independently
        searched historical evidence through $T-1$ and assigned recurring,
        new, or unresolved labels, blinded to strategy predictions, automated
        judgments, and each other's labels. A third annotator adjudicated
        disagreements remaining after evidence-based discussion.
        Because fallback review covers only reattachments initially judged
        incorrect, the blinded audit also provides an independent check on
        errors among those judged correct.
        Appendix~\ref{app:human-audit} details the evidence requirements,
        annotation agreement, metric denominators, and Wilson 95\%
        confidence intervals.

        \textbf{Human-audit results.} Compared with FIRME w/o Alias Memory,
        full FIRME increased observed human-audit CMR by 9.0--16.5 percentage
        points and reattachment recall by 22.5--29.0 points across the four
        settings (Table~\ref{tab:human-audit}). Human-audit merge precision
        was also higher in all four samples, ranging from 85.6\% to 97.5\%.
        Full FIRME achieved the highest observed human-audit CMR and recall
        among all five strategies in every setting. Its recalls on investor
        calls were 77.5\% and 68.8\% with Opus 4.6 and GPT-5.4-mini,
        respectively; corresponding S\&P 500 recalls were 88.5\% and 77.0\%
        (Table~\ref{tab:memory-usability}). RAW and spaCy-Lemma had high
        observed precision on smaller accepted subsets, with recall ranging
        from 15.0\% to 38.5\%.

        Across LLM-Norm, FIRME without Alias Memory, and Full FIRME,
        human-audit merge precision differed from population-protocol
        estimates by 0.3--5.8 percentage points, while the CMR ordering was
        unchanged in all four settings (Tables~\ref{tab:memory-usability}
        and~\ref{tab:human-audit}). The broadly similar precision estimates
        and consistent CMR ordering provide independent human evidence
        supporting the population-level finding that alias memory improves
        correct relationship recovery in the evaluated settings.

        \textbf{Token consumption.} FIRME stores accepted semantic matches
        as alias-to-canonical mappings for later reuse. Recurring aliases can
        then be resolved deterministically, whereas the no-alias control
        requires semantic fallback whenever canonical-ID and canonical-name
        matching fail.
        On approximately 1,300 investor calls in the final period $T$ using
        Opus 4.6, the three strategies have similar entity-canonicalization
        token consumption (Table~\ref{tab:token-consumption}). Compared with
        FIRME w/o Alias Memory, full FIRME reduces semantic-matching
        consumption by 40.3\% and total consumption across the two stages by
        7.2\%. These reductions are consistent with alias reuse. Measurements
        exclude earlier registry-construction costs.

        \begin{table}[tb]
            \centering
            \small
            \setlength{\tabcolsep}{3pt}
            \caption{Token consumption for approximately 1,300 investor calls
            in the final period $T$, using Opus 4.6. Totals cover entity
            canonicalization and cross-period semantic matching; earlier
            registry-construction periods are excluded. M denotes million
            tokens.}
            \label{tab:token-consumption}
            \begin{tabularx}{\columnwidth}{@{}Yrrr@{}}
                \toprule
                \textbf{Strategy} & \shortstack{\textbf{Entity}\\\textbf{canon.}} &
                \shortstack{\textbf{Semantic}\\\textbf{matching}} & \textbf{Total} \\
                \midrule
                LLM-Norm & 5.42M & --- & 5.42M \\
                FIRME w/o Alias Memory & 5.29M & 1.39M & 6.68M \\
                Full FIRME & 5.37M & 0.83M & 6.20M \\
                \bottomrule
            \end{tabularx}
        \end{table}

    \subsection{Extraction Faithfulness}
        Reliable memory accumulation requires not only cross-period identity
        continuity but also source-grounded observations within each period.
        We therefore assess extraction faithfulness through quotation
        verification, entity traceability, and entity normalization.
        Extraction faithfulness was evaluated on all final-period text-extracted
        relationships produced by each extractor--backend configuration, before
        fusion and cross-period reconciliation.
        We trace four stages:
        document, quoted evidence, raw entity mention, and normalized entity.
        \textbf{Quotation Verification Rate} is the share of extracted evidence
        texts appearing in the source, flagging fabricated quotations. The
        \textbf{Entity Traceability Rate} measures the share of raw entity mentions
        located in their associated evidence text using 0.7 token-overlap fuzzy
        matching and a language-model fallback for paraphrases. \textbf{Entity
        Normalization Accuracy} is the share of raw-to-normalized mappings judged
        semantically correct. The \textbf{Hallucination Rate} is the proportion
        of evaluated relationships whose evidence fails quotation verification
        or whose raw target fails traceability. The
        \textbf{Normalization Error Rate} equals one minus normalization
        accuracy. These diagnostics do not separately assess relation
        entailment, direction, stance, or graph completeness. Because
        verification and normalization are independent, verified edges may be
        mismapped, while correctly mapped edges may contain hallucinated
        evidence.

        GraphRAG used GPT-5.4-mini as its backend. Its text-unit identifiers
        linked schema-aligned relationships to source chunks, allowing us to
        apply the same criteria reported in Table~\ref{tab:faithfulness-comparison}.
        Appendix~\ref{app:evaluation-details} describes the adapter.
        On investor calls, Opus 4.6
        halved Hallucination Rate (3.3\% versus 6.6\%) and more than halved Normalization Error
        Rate (3.2\% versus 7.74\%) relative to GPT-5.4-mini; all Hallucination
        Rates were below 7\%. On S\&P 500, no backend dominated:
        GPT-5.4-mini lowered hallucination (5.87\% versus 7.4\%) but raised
        normalization error (22.16\% versus 2.8\%). Opus 4.6 normalization error
        was similar across corpora (3.2\% and 2.8\%). Against backend-matched GraphRAG,
        FIRME's LLM2Graph extractor with GPT-5.4-mini roughly halved hallucination
        (5.87\% versus 11.5\%) but increased normalization error
        (22.16\% versus 17.0\%), so normalization varied more than hallucination
        across LLM2Graph backends under the fixed judge.

    \begin{table}[t]
        \centering
        \small
        \setlength{\tabcolsep}{3pt}
        \caption{Extraction-faithfulness Hallucination (Hall.) and
        Normalization Error (N.Err.) rates by extractor and backend.}
        \label{tab:faithfulness-comparison}
\begin{tabular}{@{}lllrr@{}}
            \toprule
            \textbf{Dataset} & \textbf{Extractor} & \textbf{\shortstack{LLM\\Backend}} & \textbf{Hall.} & \textbf{N.Err.} \\
            \midrule
            \multirow{2}{*}{\shortstack{Investor\\calls}} & LLM2Graph & Opus 4.6  & 3.3\%  & 3.2\%  \\
                                      & LLM2Graph & GPT-5.4-mini & 6.6\%  & 7.74\% \\
            \midrule
            \multirow{3}{*}{S\&P 500} & LLM2Graph & Opus 4.6  & 7.4\%  & 2.8\%  \\
                                      & LLM2Graph & GPT-5.4-mini & 5.87\% & 22.16\% \\
                                      & GraphRAG & GPT-5.4-mini  & 11.5\% & 17.0\% \\
            \bottomrule
        \end{tabular}

    \end{table}

    \subsection{Round-Trip Recovery}
        An extractor may produce source-grounded outputs while still omitting
        information from the input during extraction. Beyond extraction
        faithfulness, we therefore evaluate preservation across representations.
        Round-trip recovery measures how much extractor-derived
        graph content survives graph-to-text generation and subsequent
        re-extraction. Each reference graph contains the complete Opus 4.6
        LLM2Graph output for one final-period call. For each corpus, we sampled
        1,000 reference graphs uniformly with replacement and generated one
        document for each sampled instance using a fixed Opus 4.6 generator.
        These documents were shared across all evaluated re-extraction backends;
        only the re-extraction backend varied. Because
        generation may add document-grounded topics absent from the reference graph,
        adjusted precision accepts an extra recovered item only when the fixed
        Gemini 3.1 Pro judge determines that the document explicitly supports it.
        We independently macro-average adjusted
        precision, recall, and F1 across instances at node and edge levels.
        Appendix~\ref{app:evaluation-details} specifies node/edge matching and
        the metric formulas.

        Table~\ref{tab:recovery} shows similar investor-call node- and edge-level
        F1 across backends (0.923--0.931). For S\&P 500, F1 is 0.862--0.865 with
        Opus 4.6 and 0.786--0.805 with GPT-5.4-mini, a larger gap. Adjusted
        precision remains 0.952--0.982, while recall carries most loss and
        reaches 0.697 on S\&P inputs. High adjusted precision means recovered
        items match reference-graph content or have explicit support; lower recall means greater
        loss of reference-graph content. With shared Opus-generated documents, S\&P-derived documents
        show lower preservation and greater backend sensitivity than
        investor-call-derived documents.

    \begin{table}[t]
        \centering
        \small
        \setlength{\tabcolsep}{2.5pt}
        \caption{Round-trip recovery: adjusted precision (Adj.\ P), recall (R),
        and F1, independently macro-averaged across instances at node and edge
        levels.}
        \label{tab:recovery}
\begin{tabular}{@{}llcrrr@{}}
            \toprule
            \textbf{Dataset} & \textbf{\shortstack{LLM\\Backend}} & \textbf{Level} & \textbf{Adj.\ P} & \textbf{R} & \textbf{F1} \\
            \midrule
            \multirow{4}{*}{\shortstack{Investor\\calls}}  & \multirow{2}{*}{Opus 4.6} & Node & 0.952 & 0.918 & 0.931 \\
                                      &                         & Edge & 0.954 & 0.917 & 0.929 \\
                                      & \multirow{2}{*}{GPT-5.4-mini} & Node & 0.963 & 0.902 & 0.926 \\
                                      &                              & Edge & 0.968 & 0.898 & 0.923 \\
            \midrule
            \multirow{4}{*}{S\&P 500} & \multirow{2}{*}{Opus 4.6} & Node & 0.982 & 0.786 & 0.865 \\
                                      &                         & Edge & 0.982 & 0.783 & 0.862 \\
                                      & \multirow{2}{*}{GPT-5.4-mini} & Node & 0.975 & 0.697 & 0.786 \\
                                      &                              & Edge & 0.973 & 0.728 & 0.805 \\
            \bottomrule
        \end{tabular}

    \end{table}

    \subsection{Human Audit of Short-Term Fusion}\label{sec:fusion-audit}
        The preceding evaluations assess identity continuity, source
        faithfulness, and content preservation. We now examine whether
        evidence from investor calls and cash-flow records supports the
        product corroborations produced by short-term fusion.

        We audited final-period investor-call data using Opus 4.6. Log records
        were deduplicated by investor, product, and period, retaining their
        evidence references. We randomly sampled 200 relationships triggering
        Rule~3 and separately sampled 300 relationships triggering Rule~3 or
        Rule~4, without replacement within each respective population.
        Within the Full sample, log tags identified the subset triggering
        Rule~4 but not Rule~3.
        Two annotators independently reviewed all sampled relationships.

        A correct corroboration required evidence from the same investor and
        period, correct product alignment, and an eligible cash-flow signal.
        Transcript evidence had to support discussion of that product for
        Rule~3, or an appropriately linked ontology topic for Rule~4. Fusion
        precision was the proportion of reviewed relationships with a valid
        evidence path under the evaluated rule set, using final judgments
        after disagreement resolution.
        Appendix~\ref{app:fusion-annotation} provides annotation instructions
        and a VTEB disagreement illustrating resolution through the rule
        criteria.

        Table~\ref{tab:fusion-audit} reports human-audit precision of 99.0\%
        for Direct fusion and 97.7\% for Full fusion. Within the Full sample,
        94 of 300 relationships (31.3\%) triggered Rule~4 without triggering
        Rule~3. Of these, 91 were judged correct, yielding a precision of
        96.8\%. Together, these findings support high precision for both
        direct and ontology-mediated corroboration in the audited setting.

        \begin{table}[tb]
            \centering
            \small
            \setlength{\tabcolsep}{3pt}
            \caption{Human audit of short-term fusion on final-period
            investor-call data using Opus 4.6. Precision uses final judgments
            after disagreement resolution. The final row is a subset of
            the 300 Full outputs and is included in the Full result.}
            \label{tab:fusion-audit}
            \begin{tabularx}{\columnwidth}{@{}Yrr@{}}
                \toprule
                \textbf{Condition / output subset} &
                \shortstack{\textbf{Audited}\\\textbf{cases}} &
                \shortstack{\textbf{Human-audit}\\\textbf{fusion precision}} \\
                \midrule
                Direct fusion (Rule~3) & 200 & 99.0\% (198/200) \\
                Full fusion (Rules~3+4) & 300 & 97.7\% (293/300) \\
                \quad Added by Rule~4 beyond Rule~3 & 94 & 96.8\% (91/94) \\
                \bottomrule
            \end{tabularx}
        \end{table}

\section{Discussion}\label{sec:discussion}

    Correct registry reattachments extend one cumulative evidence history rather
    than create duplicate \textit{Emerging} edges. FIRME's higher
    CMR indicates that a larger fraction of incoming edges is
    judged to be correctly reattached to prior memory. MMR
    describes reattachment frequency, while Merge Precision measures
    accepted-link correctness. Accepted
    registry matches add their surface forms to the selected anchor's persistent
    alias set; an incorrect registry match can therefore route later occurrences
    to the wrong anchor and history. Merge Precision therefore remains
    important alongside CMR when assessing incorrect reattachments.

    \lifecyclefigure

    \subsection{Lifecycle Behavior Over Time}
        Figure~\ref{fig:pattern} shows observed Market Volatility activity rising
        and then subsiding. Under Eqs.~\eqref{eq:sigma_update}--\eqref{eq:sigma_decay}
        and the lifecycle definitions in Section~\ref{sec:long-term-fusion},
        consecutive unmatched periods shift accumulated memory from \textit{Emerging}
        toward \textit{Stable} and \textit{Fading}; no \textit{Reactivated} share
        appears in the observed window. These states reflect observed evidence
        rather than latent investor preference.

    \subsection{Analytical Parameter Behavior}
        Table~\ref{tab:param_sensitivity} reports analytical, not empirical,
        sensitivity. Parameter $\alpha$ sets saturation speed; $\gamma$ sets the
        decay rate. Decay begins with the first miss, whereas $N$ sets a
        separate threshold for \textit{Fading}. Under the defaults, an edge starting
        at $0.5$ saturates after three consecutive matches and retains 51\% of its
        pre-decay score after three consecutive misses. Periods are monthly for
        investor calls but reporting-period-based for S\&P 500, so identical
        parameters span different calendar durations.

    \begin{table}[th]
        \centering
        \small
        \caption{Analytical effects of $\alpha$ and $\gamma$ across low,
        default, and high settings; bold marks defaults.}
        \label{tab:param_sensitivity}
\begin{tabular}{@{}lccc@{}}
            \toprule
            \textbf{Parameter / behavior} & \textbf{Low} & \textbf{Default} & \textbf{High} \\
            \midrule
            Reinforcement $\alpha$               & 0.1 & \textbf{0.2} & 0.3 \\
            \quad Matched periods to reach $1.0$ & 5   & \textbf{3}   & 2   \\
            \addlinespace
            Retention $\gamma$                   & 0.7 & \textbf{0.8} & 0.9 \\
            \quad \% retained after 3 periods    & 34  & \textbf{51}  & 73  \\
            \bottomrule
        \end{tabular}

    \end{table}

\section{Conclusion and Future Work}\label{sec:conclusion}

    We introduced FIRME, a registry-anchored memory-graph framework for
    heterogeneous, time-varying investor evidence. Within-period ontology-mediated fusion
    combines document-derived and structured evidence with provenance.
    Cross-period registry reconciliation maps surface variants to canonical
    relationships, allowing evidence to accumulate in one history through
    deterministic lifecycle updates. Under the judge and fallback-review
    protocol, FIRME increased CMR over the no-alias control by 7.8--12.1
    percentage points and achieved 2.3--3.7 times LLM-Norm's CMR.
    The shared human audit found CMR gains of 9.0--16.5 points and
    reattachment-recall gains of 22.5--29.0 points over the no-alias control,
    with higher observed merge precision in all four samples. Results support persistent
    identity for auditable evidence trajectories, not latent investor preference.

    Future work should learn entity- and relation-specific decay, incorporate
    negative signals (outflows and dissatisfaction), support firm-level
    aggregation, and test longer windows, more backends, and independent judges
    and generators.

\section{Limitations}
    FIRME's lifecycle model is implemented as a deterministic state machine.
    Given reconciliation decisions, its memory-strength score and lifecycle
    state follow fixed, evidence-grounded updates; extraction and semantic
    matching remain model-dependent.
    The alias-memory ablation evaluates persistent aliases as a combined
    mechanism, but does not separately quantify the effects of canonical
    registry state and semantic fallback.

    The proprietary investor-call corpus and the internal product-annotation
    resource underlying the ontology are not publicly available, limiting
    independent reproduction of the corresponding experiments. The S\&P 500 corpus
    contains transcripts for 297 sampled companies across three reporting
    periods but no paired structured cash-flow records. It therefore tests
    extraction and temporal reconciliation, not the full heterogeneous-fusion
    path. Neither window spans an annual cycle, limiting longer-term
    seasonality and lifecycle conclusions.

    The human audit estimates reattachment recall for sampled incoming
    units against an adjudicated historical reference; a complete population
    reference remains unavailable. It is conditional on each backend's
    extracted observation stream and does not measure extraction recall or
    relationships omitted during extraction from the original calls.
    Estimates remain subject to sampling uncertainty and errors in historical
    search or reference labeling. The confidence intervals quantify sampling
    uncertainty conditional on the final reference.

    The product--topic ontology, \$100K threshold, and lifecycle parameters are
    use-case specific; uniform reinforcement ignores relation type and evidence
    intensity. At memory update, an edge is unmatched when no current observation
    is successfully extracted, registry-normalized, and linked to it. This may
    reflect reduced engagement, missing inputs, or no observation opportunity,
    which the model cannot distinguish.

    Two backends limit model-level generalization. Using Gemini 3.1 Pro as the
    fixed judge keeps the evaluator constant while the extractor varies.
    The manual fallback review and human audit may retain
    annotator subjectivity. The Gemini 3.1 Pro judge is separate from the
    Opus 4.6 extractor, although reliance on a single LLM judge may introduce
    evaluator-specific bias. Opus 4.6 also generated all round-trip documents;
    sharing them controls input variation, although generator--extractor
    affinity may remain.

\nobalance
\bibliographystyle{ACM-Reference-Format}
\bibliography{graph}

\clearpage
\nobalance
\onecolumn
\appendix
\section*{\Large Appendix}
\par\addvspace{6pt}
\numberwithin{table}{section}
\renewcommand{\thetable}{\thesection\arabic{table}}
\ifdefined\theHtable
    \renewcommand{\theHtable}{\thesection.\arabic{table}}
\fi
\section{Human Audit of Relationship Reattachment}\label{app:human-audit}

\begingroup
\setlength{\multicolsep}{4pt}
\setlength{\columnsep}{1cm}
\begin{multicols}{2}
\raggedcolumns
\paragraph{Sampling and historical reference.}
For each dataset--backend setting, we sampled 200 final-period incoming
relationship units uniformly without replacement from the complete incoming
population, including units left unmatched by the strategies. Each incoming
observation counted as one unit; observations were not deduplicated for
evaluation. Multiple observations could originate from the same call.
All five strategies shared
the sampled units and reference within a setting. RAW and spaCy-Lemma
predictions were scored against the existing frozen reference without
repeating annotation. Extracted streams and sampled units could differ
between Opus 4.6 and GPT-5.4-mini.

Two annotators independently reviewed all 800 units, yielding 1,600 initial
reviews. Each of the two primary annotators spent approximately 65 hours
on annotation. They inspected incoming source evidence and searched historical
relationship observations and supporting evidence available through $T-1$.
The eligible history and identity criteria were shared across strategies.
Historical search extended beyond strategy-proposed candidates. Annotators
were blinded to strategy predictions, automated judgments, and each other's
initial labels.

\paragraph{Evidence-aware labels.}
Each unit received a MATCH, NEW, or UNRESOLVED label. MATCH required an
evidence-supported historical relationship with the same head and typed target
identity. Supported aliases and synonymous expressions could identify the
same target; relationship-label differences alone did not establish different
identities. NEW required a search of eligible history that found no
corresponding relationship. Insufficient evidence led to UNRESOLVED.
Annotation records linked incoming units and historical counterparts to
witness edges, call identifiers, turn identifiers or stable source locators,
supporting excerpts, and a short rationale.

\paragraph{Independent agreement and adjudication.}
Initial labels were preserved before discussion. Substantive agreement
required both annotators to select NEW or MATCH to the same historical
identity. Different valid evidence passages supporting that identity counted
as agreement. Different decisions or identities, initial uncertainty, and
insufficient supporting evidence entered resolution. The two annotators
discussed these cases using their evidence records; a third annotator
adjudicated remaining disagreements. Agreement was computed from independent
initial annotations, and strategy performance used the final adjudicated
reference. Initial substantive agreement was 704/800 (88.0\%); 22 of the
96 cases requiring resolution received third-person review. All final
labels were resolved (Table~\ref{tab:human-audit-agreement}).
\end{multicols}
\par\medskip
\noindent\begin{minipage}{\textwidth}
\centering
\small
\renewcommand{\arraystretch}{1.15}
\setlength{\tabcolsep}{4pt}
\captionsetup{type=table,skip=5pt}
\caption{Human-audit annotation agreement and reference composition.
Each setting contains 200 units. Agreement precedes discussion;
third-person reviews are a subset of cases requiring resolution.}
\label{tab:human-audit-agreement}
\begin{tabularx}{\textwidth}{@{}l*{4}{>{\centering\arraybackslash}X}@{}}
\toprule
\textbf{Setting} & \textbf{\shortstack{Initial substantive\\agreement}} &
\textbf{\shortstack{Cases requiring\\resolution}} &
\textbf{\shortstack{Third-person\\reviews}} &
\textbf{\shortstack{Final recurring /\\new / unresolved}} \\
\midrule
Investor $\times$ Opus 4.6 & 176/200 (88.0\%) & 24 & 6 & 80 / 120 / 0 \\
Investor $\times$ GPT-5.4-mini & 172/200 (86.0\%) & 28 & 7 & 80 / 120 / 0 \\
S\&P 500 $\times$ Opus 4.6 & 182/200 (91.0\%) & 18 & 4 & 130 / 70 / 0 \\
S\&P 500 $\times$ GPT-5.4-mini & 174/200 (87.0\%) & 26 & 5 & 100 / 100 / 0 \\
\midrule
\textbf{Total} & \textbf{704/800 (88.0\%)} & \textbf{96} & \textbf{22} & \textbf{390 / 410 / 0} \\
\bottomrule
\end{tabularx}
\end{minipage}
\endgroup

\clearpage
\begingroup
\setlength{\multicolsep}{4pt}
\setlength{\columnsep}{1cm}
\begin{multicols}{2}
\raggedcolumns
\paragraph{Performance measures and uncertainty.}
A reattachment was correct when its historical relationship identity matched
the adjudicated reference. Human-audit CMR divides correct reattachments by
all 200 sampled incoming units. Human-audit merge precision divides by
accepted reattachments; human-audit reattachment recall divides by gold
recurring units. A recurring unit attached to the wrong historical identity
counts as both an incorrect accepted reattachment and a failure to recover
the correct relationship. A new unit correctly left unmatched does not count
as a correct reattachment. Human-audit F1 is the harmonic mean of
precision and recall, $2PR/(P+R)$, using unrounded values from the same
adjudicated reference.

Table~\ref{tab:human-audit} reports CMR, precision, and recall with two-sided
Wilson 95\% confidence intervals using each metric's denominator; F1 is
reported as a point estimate. These
pointwise intervals use a binomial approximation to random incoming-unit
sampling without finite-population correction, conditional on the adjudicated
reference. No units were excluded for unresolved final labels. Reattachment
recall is conditional on the extracted streams and eligible history;
extraction recall over the original transcripts is outside this audit's scope.

\paragraph{Connection to Table~\ref{tab:memory-usability}.}
RAW and spaCy-Lemma human-audit precision also supplies their Table~\ref{tab:memory-usability}
precision entries. Their CMR there multiplies the reported population MMR
by this precision; Table~\ref{tab:human-audit} computes human-audit CMR
directly as correct reattachments divided by 200. The former uses population
acceptance rates and the latter uses sample acceptance rates, so their CMR
estimates can differ. Baseline precision estimates use 12--51 accepted
cases per setting, with uncertainty reported below.

\paragraph{F1 comparison.}
Full FIRME achieved the highest observed human-audit F1 in every setting
(76.4--92.7\%), exceeding FIRME w/o Alias Memory by 17.2--20.5 percentage
points.
\end{multicols}
\par\medskip
\noindent\begin{minipage}{\textwidth}
\centering
\fontsize{8.5}{10.5}\selectfont
\setlength{\tabcolsep}{2pt}
\renewcommand{\arraystretch}{1.18}
\captionsetup{type=table,skip=6pt}
\caption{Human-audit performance on 200 shared incoming units per setting.
Gold recurring counts are shared across strategies; accepted and correct
reattachments are strategy-specific. All metrics are percentages; brackets
give Wilson 95\% confidence intervals for CMR, precision, and recall. F1 is
a point estimate. Bold marks the highest CMR, recall, and F1 within each setting.}
\label{tab:human-audit}
\begin{tabularx}{\textwidth}{@{}llrrr*{3}{>{\centering\arraybackslash}X}r@{}}
\toprule
\textbf{\shortstack{Dataset $\times$\\backend}} & \textbf{Strategy} &
\textbf{\shortstack{Gold\\recurring}} & \textbf{\shortstack{Accepted\\reattach.}} &
\textbf{\shortstack{Correct\\reattach.}} &
\textbf{\shortstack{Human-audit\\CMR}} &
\textbf{\shortstack{Human-audit\\merge precision}} &
\textbf{\shortstack{Human-audit\\reattachment recall}} & \textbf{F1} \\
\midrule
\multirow{5}{*}{\shortstack[l]{Investor $\times$\\Opus 4.6}} & RAW & 80 & 12 & 12 & 6.0 [3.5, 10.2] & 100.0 [75.8, 100.0] & 15.0 [8.8, 24.4] & 26.1 \\
 & spaCy-Lemma & 80 & 17 & 16 & 8.0 [5.0, 12.6] & 94.1 [73.0, 99.0] & 20.0 [12.7, 30.0] & 33.0 \\
 & LLM-Norm & 80 & 17 & 16 & 8.0 [5.0, 12.6] & 94.1 [73.0, 99.0] & 20.0 [12.7, 30.0] & 33.0 \\
 & \shortstack[l]{FIRME w/o\\Alias Memory} & 80 & 48 & 42 & 21.0 [15.9, 27.2] & 87.5 [75.3, 94.1] & 52.5 [41.7, 63.1] & 65.6 \\
 & \textbf{Full FIRME} & 80 & 64 & 62 & \textbf{31.0 [25.0, 37.7]} & 96.9 [89.3, 99.1] & \textbf{77.5 [67.2, 85.3]} & \textbf{86.1} \\
\midrule
\multirow{5}{*}{\shortstack[l]{Investor $\times$\\GPT-5.4-mini}} & RAW & 80 & 12 & 12 & 6.0 [3.5, 10.2] & 100.0 [75.8, 100.0] & 15.0 [8.8, 24.4] & 26.1 \\
 & spaCy-Lemma & 80 & 13 & 13 & 6.5 [3.8, 10.8] & 100.0 [77.2, 100.0] & 16.3 [9.7, 25.8] & 28.0 \\
 & LLM-Norm & 80 & 18 & 16 & 8.0 [5.0, 12.6] & 88.9 [67.2, 96.9] & 20.0 [12.7, 30.0] & 32.7 \\
 & \shortstack[l]{FIRME w/o\\Alias Memory} & 80 & 45 & 37 & 18.5 [13.7, 24.5] & 82.2 [68.7, 90.7] & 46.3 [35.7, 57.1] & 59.2 \\
 & \textbf{Full FIRME} & 80 & 64 & 55 & \textbf{27.5 [21.8, 34.1]} & 85.9 [75.4, 92.4] & \textbf{68.8 [57.9, 77.8]} & \textbf{76.4} \\
\midrule
\multirow{5}{*}{\shortstack[l]{S\&P 500 $\times$\\Opus 4.6}} & RAW & 130 & 28 & 28 & 14.0 [9.9, 19.5] & 100.0 [87.9, 100.0] & 21.5 [15.3, 29.4] & 35.4 \\
 & spaCy-Lemma & 130 & 51 & 50 & 25.0 [19.5, 31.4] & 98.0 [89.7, 99.7] & 38.5 [30.5, 47.0] & 55.2 \\
 & LLM-Norm & 130 & 48 & 47 & 23.5 [18.2, 29.8] & 97.9 [89.1, 99.6] & 36.2 [28.4, 44.7] & 52.8 \\
 & \shortstack[l]{FIRME w/o\\Alias Memory} & 130 & 88 & 82 & 41.0 [34.4, 47.9] & 93.2 [85.9, 96.8] & 63.1 [54.5, 70.9] & 75.2 \\
 & \textbf{Full FIRME} & 130 & 118 & 115 & \textbf{57.5 [50.6, 64.1]} & 97.5 [92.8, 99.1] & \textbf{88.5 [81.8, 92.9]} & \textbf{92.7} \\
\midrule
\multirow{5}{*}{\shortstack[l]{S\&P 500 $\times$\\GPT-5.4-mini}} & RAW & 100 & 21 & 21 & 10.5 [7.0, 15.5] & 100.0 [84.5, 100.0] & 21.0 [14.2, 30.0] & 34.7 \\
 & spaCy-Lemma & 100 & 33 & 32 & 16.0 [11.6, 21.7] & 97.0 [84.7, 99.5] & 32.0 [23.7, 41.7] & 48.1 \\
 & LLM-Norm & 100 & 32 & 30 & 15.0 [10.7, 20.6] & 93.8 [79.9, 98.3] & 30.0 [21.9, 39.6] & 45.5 \\
 & \shortstack[l]{FIRME w/o\\Alias Memory} & 100 & 57 & 48 & 24.0 [18.6, 30.4] & 84.2 [72.6, 91.5] & 48.0 [38.5, 57.7] & 61.1 \\
 & \textbf{Full FIRME} & 100 & 90 & 77 & \textbf{38.5 [32.0, 45.4]} & 85.6 [76.8, 91.4] & \textbf{77.0 [67.8, 84.2]} & \textbf{81.1} \\
\bottomrule
\end{tabularx}
\end{minipage}
\endgroup

\clearpage
\section{Reconciliation Strategies and Alias-Memory Control}
\label{app:alias-memory-control}

Table~\ref{tab:strategy-roles} summarizes the three strategies used to
interpret the contribution of persistent alias memory. FIRME w/o Alias
Memory retains historical canonical identities and cross-period semantic
fallback, so its comparison with full FIRME evaluates the incremental
contribution of alias memory beyond canonical-only historical reconciliation.
\par\medskip

\begingroup
\small
\setlength{\tabcolsep}{4pt}
\renewcommand{\arraystretch}{1.2}
\captionsetup{type=table}
\caption{Reconciliation capabilities and comparison roles. Historical
registry use refers to resolving incoming names against existing canonical
identities; LLM-Norm can still reattach equal normalized names or identifiers
through deterministic matching.}
\label{tab:strategy-roles}
\begin{tabularx}{\textwidth}{@{}>{\raggedright\arraybackslash}p{0.14\textwidth}>{\raggedright\arraybackslash}p{0.18\textwidth}>{\raggedright\arraybackslash}p{0.175\textwidth}>{\raggedright\arraybackslash}p{0.10\textwidth}>{\raggedright\arraybackslash}p{0.115\textwidth}Y@{}}
    \toprule
    \textbf{Strategy} & \textbf{Brief description} &
    \textbf{Historical registry used during reconciliation} &
    \textbf{Learned alias memory} & \textbf{Cross-period semantic fallback} &
    \textbf{Role in the comparison} \\
    \midrule
    LLM-Norm & Independent within-period semantic normalization & No & No & No &
    Reference for independent normalization \\
    \addlinespace
    FIRME w/o Alias Memory & Canonical registry with deterministic and semantic matching, without alias memory &
    Yes; canonical IDs and names & No & Yes &
    Controls for historical reconciliation and semantic fallback while removing alias memory \\
    \addlinespace
    Full FIRME & Registry reconciliation with persistent alias reuse &
    Yes; canonical IDs, names, and aliases & Yes & Yes &
    Evaluates the incremental contribution of alias memory relative to the no-alias control \\
    \bottomrule
\end{tabularx}
\par
\endgroup

The no-alias control disables learned-alias accumulation, deterministic alias
lookup, and alias information in semantic candidate descriptions. Both FIRME
variants retain cross-period semantic fallback and maintain their own evolving
registries over the same period sequence, with the same incoming observations,
backend, matching threshold, and fusion rules. The no-alias registry retains
accepted canonical assignments but does not accumulate their alternative
names. LLM-Norm can still reattach equal normalized names or identifiers
through deterministic matching; it does not use historical canonical
identities to resolve incoming names.

\twocolumn
\raggedbottom
\section{Cross-Period Entity Reconciliation Pseudocode}
\label{app:entity-reconciliation}

The procedure below updates the entity registry for $t\geq1$, after within-period
fusion. The first period establishes the registry; subsequent periods reconcile
incoming entities against the prior registry. Candidates include all original
target entities retained in $S^t$, including topic antecedents used by Rule~4.

\noindent\textbf{Initialization:} Set $R^0=\varnothing$ before processing
the first period.

\noindent\textbf{Input:} Incoming graph $S^t$, prior registry $R^{t-1}$.

\noindent\textbf{Output:} Updated entity registry $R^t$.

\begin{center}
\small
\renewcommand{\arraystretch}{1.12}
\begin{tabularx}{\columnwidth}{@{}r@{\hspace{0.5em}}X@{}}
\toprule
1 & \textbf{if} $t=1$: \\
2 & \quad Initialize $R^1$ from deduplicated target entities in $S^1$,
    retaining their aliases \\
3 & \quad \textbf{return} $R^1$ \\
4 & $R^t \gets \operatorname{copy}(R^{t-1})$ \\
5 & \textbf{for each} distinct target entity $v$ in $S^t$: \\
6 & \quad $\mathit{anchor} \gets \operatorname{deterministic\_match}(v,R^{t-1})$ \\
7 & \quad \textbf{if} no anchor is found: \\
8 & \qquad $\mathit{anchor} \gets \operatorname{semantic\_match}(v,R^{t-1})$ \\
9 & \quad \textbf{if} an anchor is found: \\
10 & \qquad Add $v$'s surface forms to the anchor's alias set in $R^t$ \\
11 & \quad \textbf{else}: \\
12 & \qquad Register $v$ with its original identifier in $R^t$ \\
13 & \textbf{return} $R^t$ \\
\bottomrule
\end{tabularx}
\end{center}

At $t=1$, deduplication uses lemmatized-name equality, followed by semantic
matching for unresolved same-type candidates, as described in
Section~\ref{sec:problem-statement}. For $t>1$, deterministic matching checks identifiers,
normalized names, and learned aliases in sequence, stopping at the first match.
Semantic matching considers canonical names and aliases of same-type entities
in the prior registry. All matching queries for these subsequent periods read
$R^{t-1}$; new identities and accepted aliases are written to $R^t$.
Matched incoming targets adopt the selected canonical IDs for edge
reconciliation; head IDs remain unchanged.

\newpage
\section{Exploratory Semantic Threshold Selection}
\label{app:semantic-threshold}


\paragraph{Evaluation setup.}
We evaluated all 5,610 distinct entities that remained unresolved after
deterministic identifier, normalized-name, and alias matching in the penultimate
period ($T-1$) of the investor-call corpus. The same entity set was evaluated at
thresholds of 0.7, 0.8, and 0.9 using Opus 4.6. The LLM scored same-type
registry candidates using their canonical names and aliases, following a
similarity rubric with worked examples. At each threshold, the rule-based
matcher selected the highest-scoring candidate whose score met or exceeded
the threshold.

\paragraph{Entity match rate.}
At each threshold, we calculate the fraction of evaluated unresolved entities
accepted by semantic matching:
\begingroup
\fontsize{10}{12}\selectfont
\[
    \text{Entity match rate} =
    \frac{\text{\# matched entities}}
         {\text{\# evaluated unresolved entities}}.
\]
\endgroup

\paragraph{Human-audited precision.}
Two annotators independently reviewed accepted semantic matches, each
assessing 100 matches per threshold. A match was judged correct when the
incoming entity and the selected registry identity denoted the same entity.
Each annotator's precision was the number of correct matches divided by 100.
We report the arithmetic mean of the two annotators' precision estimates.

\par\addvspace{8pt}
\noindent\begin{minipage}{\columnwidth}
    \centering
    \small
    \renewcommand{\arraystretch}{1.15}
    \setlength{\tabcolsep}{3pt}
    \captionsetup{type=table,skip=6pt}
    \caption{Exploratory threshold analysis for semantic entity matching on
    investor-call data using Opus 4.6. Bold indicates the selected threshold.}
    \label{tab:semantic-threshold}
\begin{tabularx}{\columnwidth}{@{}c*{2}{>{\centering\arraybackslash}X}@{}}
        \toprule
        \textbf{Threshold} & \textbf{Entity match rate} &
        \textbf{Mean human-audited precision} \\
        \midrule
        0.7 & 41.7\% & 85.0\% \\
        \textbf{0.8} & \textbf{35.5\%} & \textbf{94.5\%} \\
        0.9 & 17.0\% & 95.5\% \\
        \bottomrule
    \end{tabularx}

\end{minipage}
\par\addvspace{6pt}

\paragraph{Threshold selection.}
We selected 0.8 to improve precision over 0.7 while retaining more matches
than 0.9 (Table~\ref{tab:semantic-threshold}).

\onecolumn
\section{Rule-Based Fusion and Memory Continuity}
\label{app:fusion-memory-examples}

\subsection{Rule-Based Short-Term Fusion}
\label{app:short-term-fusion}

The procedure summarizes all four rules in Table~\ref{tab:fusion_rules} for one
period. Product identities are aligned as in Section~\ref{sec:short-term-fusion},
and $G_s^t$ contains only eligible cash-flow signals.
Using the paper's $(u,r,v)$ notation, $u$ denotes the investor (head), and $v$
denotes the product target in the cash-flow edges below.

\noindent\textbf{Input:} $G_u^t$, $G_s^t$, and ontology $O$.
\quad\textbf{Output:} Fused graph $S^t$.

\begin{center}
\small
\renewcommand{\arraystretch}{1.12}
\begin{tabularx}{\textwidth}{@{}r@{\hspace{0.5em}}X@{}}
\toprule
1 & $S^t \gets \operatorname{retain\_and\_group}(G_u^t \cup G_s^t)$ \\
2 & \textbf{for each} distinct edge $(u,\texttt{cf-signal},v)\in G_s^t$: \\
3 & \quad \textbf{if} Rule~3 or~4 applies to this edge: \\
4 & \qquad Label the relationship from $u$ to $v$ in $S^t$ as \texttt{strong-interest}, source~3 \\
5 & \qquad Reference all matching observations and applicable ontology links \\
6 & \textbf{return} $S^t$ \\
\bottomrule
\end{tabularx}
\end{center}

Initialization groups relationships by investor and typed target, retaining
each original observation once with its target, relation, source, and
provenance. All matches of Rules~3--4 are evaluated against the unchanged current-period inputs;
upgrades affect only the derived product label and source indicator.
Supporting observations are referenced without being consumed or copied.
Topic relationships and uncorroborated product relationships retain their
original labels.

\subsection{Topic and Product Histories Across Periods}

Emma discusses Tax Efficiency in three successive periods. In the second
period, she also has an eligible cash-flow signal for VTEB, which the ontology
links to Tax Efficiency. Rule~4 assigns \texttt{strong-interest} to the VTEB
relationship while retaining \texttt{discussed} for the topic.
Table~\ref{tab:fusion-continuity} illustrates the deterministic memory updates;
it is not a measured investor trajectory.

\par\addvspace{6pt}
\noindent\begin{minipage}{\textwidth}
    \centering
    \small
    \setlength{\tabcolsep}{5pt}
    \renewcommand{\arraystretch}{1.15}
    \captionsetup{type=table,skip=6pt}
    \caption{Illustrative continuity of Emma's topic and product histories.
    Memory columns report score and lifecycle state. Each relationship receives
    one update per period; a new relationship is initialized without an
    additional reinforcement.}
    \label{tab:fusion-continuity}
    \begin{tabularx}{\textwidth}{@{}cYll@{}}
        \toprule
        \textbf{Period} & \textbf{Current observations} &
        \textbf{Topic memory} & \textbf{VTEB memory} \\
        \midrule
        1 & Topic Obs.~1: Tax Efficiency discussion
          & $0.50$, Emerging & Not yet present \\
        2 & Topic Obs.~2: Tax Efficiency discussion;\newline
            Cash-flow Obs.~1: VTEB inflow
          & $0.70$, Reinforced & $0.50$, Emerging \\
        3 & Topic Obs.~3: Tax Efficiency discussion
          & $0.90$, Reinforced & $0.40$, Stable \\
        \bottomrule
    \end{tabularx}
\end{minipage}
\par\addvspace{6pt}

The topic history retains all three discussion observations. The VTEB history
retains the cash-flow observation and a reference to the supporting discussion
in Period~2. In Period~3, the topic is reinforced, while VTEB decays because it
has no new observation. Historical support remains retrievable but does not
trigger another update.

If one discussion corroborates several allocated products, each product
references the same discussion. The topic history and each observed product
history are updated once per period, regardless of the number of support
paths. This also applies when Rules~3 and~4 both support the same product.
An ontology link alone creates no topic observation: without a current topic
discussion, a cash-flow signal does not trigger Rule~4 or reinforce that topic.

\par\addvspace{12pt}
\noindent\begin{minipage}{\textwidth}
    \centering
    \small
    \setlength{\tabcolsep}{9pt}
    \renewcommand{\arraystretch}{1.12}
    \captionsetup{type=table,skip=6pt}
    \caption{Illustrative memory-strength updates with initialization $0.50$,
    reinforcement $0.20$, retention $0.80$, and inactivity threshold $N=3$.
    Step~0 denotes this relationship's initialization, not global period~0.
    Reinforcement is capped at $1.00$.}
    \label{tab:status-evolution}
    \begin{tabular}{@{}cllcl@{}}
        \toprule
        \textbf{Step} & \textbf{Event} & \textbf{Calculation} &
        \textbf{Score} & \textbf{Memory state} \\
        \midrule
        0 & New       & Initialization                   & 0.50 & Emerging \\
        1 & Matched   & $0.50+0.20$                      & 0.70 & Reinforced \\
        2 & Matched   & $0.70+0.20$                      & 0.90 & Reinforced \\
        3 & Matched   & $\min(1.00,\,0.90+0.20)$          & 1.00 & Reinforced \\
        4 & Unmatched & $1.00\times0.80$                 & 0.80 & Stable \\
        5 & Unmatched & $0.80\times0.80$                 & 0.64 & Stable \\
        6 & Matched   & $0.64+0.20$                      & 0.84 & Reinforced \\
        \bottomrule
    \end{tabular}
\end{minipage}

\clearpage
\section{System Prompts}
\label{app:prompt-templates}
\subsection{Relationship Extraction}
\label{app:relationship-extraction-prompt}

Both investor-call prompt listings replace legacy advisor wording and role
labels with investor, referring to the same business entity and identifier.
The extraction prompt below describes the investor-call schema; the S\&P 500
extraction uses a broader entity-type schema.

\noindent\begin{minipage}{\textwidth}
\begin{lstlisting}[style=prompttemplate,
    basicstyle=\fontsize{8.7}{9.1}\selectfont\ttfamily,
    framesep=6pt,xleftmargin=6.4pt,xrightmargin=6.4pt,aboveskip=6pt]
You are a financial domain relationship extractor. Extract relationships between an investor and products/topics substantively discussed in a call.

For each extracted relationship, return two linked layers:
- Raw: source evidence and mention metadata.
- Canonical: normalized fields for graph construction.

Raw layer:
- raw_subject: the speaker who mentioned the entity.
- raw_relation: the exact relation phrase from the transcript.
- raw_object: the exact entity mention, copied unchanged
  from the transcript.
- speaker_role: e.g. "investor" or "crew".
- mention_origin: e.g. "investor_proactive", "investor_responsive",
  or "crew_introduced".
- evidence_text: a contiguous, verbatim transcript excerpt
  containing raw_object.
- turn_ids: the source turn IDs.
- confidence: extraction confidence between 0.0 and 1.0.

Canonical layer:
- subject_entity_id: the investor's stable entity ID, regardless of speaker.
- relation_type: "DISCUSSED".
- object_entity_name: the normalized/canonical entity name.
- object_entity_type: "product" or "topic".
- entity_match_status: "known" or "unknown".

Rules:
- Extract only from the transcript. Do not infer relationships
  from general knowledge.
- Preserve raw_relation, raw_object, and evidence_text verbatim.
  Do not correct grammar, expand abbreviations, or reword them.
- Unknown entities should still be extracted.
- Return valid JSON only.

Output format:
{
  "relationships": [
    {
      "raw": {
        "raw_subject": "...",
        "raw_relation": "...",
        "raw_object": "...",
        "speaker_role": "investor",
        "mention_origin": "investor_proactive",
        "evidence_text": "...",
        "turn_ids": [1],
        "confidence": 0.0
      },
      "canonical": {
        "subject_entity_id": "...",
        "relation_type": "DISCUSSED",
        "object_entity_name": "...",
        "object_entity_type": "topic",
        "entity_match_status": "known"
      }
    }
  ]
}
\end{lstlisting}
\end{minipage}

\clearpage
\subsection{Semantic Entity Matching}
\label{app:semantic-entity-matching-prompt}

The following system prompt was used for semantic entity matching by full
FIRME in the investor-call experiments. A similar prompt with minor adaptations
was used for the S\&P 500 experiments. It is applied only to incoming entities
that remain unresolved after deterministic matching. The LLM scores all supplied
same-type registry candidates using their canonical names and aliases.
For FIRME w/o Alias Memory, candidate descriptions omit alias information,
as specified in Appendix~\ref{app:alias-memory-control}. Thresholding,
highest-score selection, and canonical-ID assignment are performed
programmatically as described in Section~\ref{sec:long-term-fusion}.

\noindent\begin{minipage}{\textwidth}
\begin{lstlisting}[style=prompttemplate]
You are a knowledge graph entity matcher for investor interest tracking.

Input:
You are given a batch of short-term edges and their associated same-type registry candidates. Each registry candidate includes a canonical ID, a canonical name, and aliases.

For each incoming target entity from the short-term edges, score its semantic similarity to every supplied same-type registry candidate using the candidate's canonical name and aliases.

Scoring:
- 1.0: Same entity or concept.
- 0.8 <= score < 1.0: Likely the same entity or concept.
- 0.5 <= score < 0.8: Related but distinct.
- 0.0 <= score < 0.5: Different meanings.

Examples:
- "Municipal Bonds" and "Muni Bond": equivalent entities (0.9+).
- "Tax-Free product" and "Tax-Exempt product": closely related concepts (0.8+).
- "Fixed Income" and "Growth Equities": distinct concepts (0.2).

Output format:
Return only JSON keyed by incoming entity name. Each value must list all supplied candidates, with one numeric score per candidate. Always return the candidate's canonical name and canonical ID exactly as provided, including when similarity is established through an alias.

Example output:
{
  "Municipal Bonds": [
    {
      "memory_entity": "Muni Bond",
      "memory_entity_id": "XXXX",
      "similarity_score": 1.0
    }
  ]
}
\end{lstlisting}
\end{minipage}

\clearpage
\section{Product--Topic Ontology: Representation and Use}
\label{app:ontology-examples}

The ontology is derived from a proprietary internal resource manually
annotated by our financial analyst team and covering approximately 17,000
unique financial products. This resource is not publicly available.
For the investor-call use case, we extracted the relevant product--topic
associations from this broader resource; selected records are shown below.

We use a fixed product--topic ontology $O$ as background domain knowledge
for short-term fusion. In this work, $O$ denotes a lightweight set of
associations between financial products and investment topics. A link
$(p,t)\in O$ indicates that product $p$ is associated with topic $t$.
A product may be associated with multiple topics, and a topic may be
associated with multiple products. These associations express topical
relevance rather than entity equivalence.

\noindent\begin{minipage}{\textwidth}
\begin{lstlisting}[style=prompttemplate]
[
  {
    "product_id": "VFIAX",
    "product_name": "Vanguard 500 Index Fund Admiral Shares",
    "product_category": "Large Blend",
    "topic_name": [
      "Passive / Index Investing",
      "Large Cap"
    ],
    "relation": "RELATED_TO"
  },
  {
    "product_id": "VTIP",
    "product_name": "Vanguard Short-Term Inflation-Protected Securities ETF",
    "product_category": "Short-Term Inflation-Protected Bond",
    "topic_name": [
      "Fixed Income",
      "Inflation-Protected Bond / TIPS",
      "Short Duration / Short End of Curve"
    ],
    "relation": "RELATED_TO"
  },
  {
    "product_id": "VGT",
    "product_name": "Vanguard Information Technology ETF",
    "product_category": "Technology",
    "topic_name": [
      "Tech Stocks",
      "Sector Funds",
      "Passive / Index Investing"
    ],
    "relation": "RELATED_TO"
  }
]
\end{lstlisting}
\end{minipage}
\par\addvspace{6pt}
\noindent
Under Rule~4, these associations connect topic-level discussion with
product-level cash-flow evidence. For example, suppose an investor discusses
Large Cap and has an eligible cash-flow signal for VFIAX in the same period.
Because $(\text{VFIAX},\text{Large Cap})\in O$, the discussion corroborates
the product allocation. FIRME labels the investor--VFIAX relationship as
\texttt{strong-interest} while retaining the investor--Large Cap relationship
as \texttt{discussed}. An ontology link alone does not constitute an investor
observation or trigger this upgrade.

\clearpage
\section{Additional Evaluation Details}
\label{app:evaluation-details}
\begingroup
\setlength{\multicolsep}{4pt}
\setlength{\columnsep}{1cm}
\begin{multicols}{2}
\raggedcolumns
\paragraph{Population judge and human fallback.}
For each accepted reattachment, the judge received the incoming relationship
and its source evidence, together with the selected historical relationship
and supporting evidence available through $T-1$. Correctness required the
same head and typed target identity. Two annotators reviewed cases initially
judged incorrect; disagreements were resolved through evidence-based
discussion, and the resulting human label replaced the initial judgment.

\paragraph{GraphRAG adapter.}
The GraphRAG adapter retained relationship-level raw mentions, normalized
targets, and candidate evidence excerpts from the extraction outputs.
Text-unit identifiers located the original source passages, against which
the candidate evidence and mentions were checked. Normalization was assessed
on the corresponding raw-to-normalized target pairs.

\paragraph{Round-trip matching and scoring.}
Nodes were matched one-to-one using entity type and normalized-name equality
under a fixed normalization rule. An edge matched only when its aligned
endpoints, relation type, and direction matched the reference edge.
For each sampled instance, let $n_{\mathrm{ref}}$ and $n_{\mathrm{rec}}$
count reference and recovered items, respectively, and let
$n_{\mathrm{match}}$ count matched items. Let $n_{\mathrm{extra}}$ count
otherwise unmatched recovered items judged by Gemini 3.1 Pro to be explicitly
supported by the generated document. At each of the node and edge levels,
\begin{align*}
P_{\mathrm{adj}} &= \frac{n_{\mathrm{match}}+n_{\mathrm{extra}}}
                         {n_{\mathrm{rec}}}, \\
R &= \frac{n_{\mathrm{match}}}{n_{\mathrm{ref}}}, \\
F_1 &= \frac{2P_{\mathrm{adj}}R}{P_{\mathrm{adj}}+R}.
\end{align*}
Grounded extra items contribute to adjusted precision but do not count as
recovered reference items. We calculate each metric per sampled instance
and then macro-average it separately across instances.

\subsection{Short-Term Fusion Annotation}\label{app:fusion-annotation}
\paragraph{Annotation instructions.}
The review unit was an investor--product--period relationship, together with
its retained transcript evidence, structured cash-flow records, and any
cited product--topic ontology entry. Reviewers checked three aspects:
investor, product, and period alignment; source support for the discussion
and eligible cash-flow signal; and justification under the applicable fusion
rule. Rule~3 required discussion of the same product as the cash-flow signal.
Rule~4 required discussion of a topic appropriately associated with that
product in the frozen ontology. Direct outputs required a valid Rule~3 path;
Full outputs required at least one valid Rule~3 or Rule~4 path. Correctness
therefore concerned the evidence supporting the corroborated relationship.

Two annotators independently assessed every relationship in both samples.
The reported counts reflect their final judgments after disagreement
resolution.

\paragraph{Disagreement resolution: a VTEB case.}
One disagreement concerned an investor--VTEB relationship. Annotator~A judged
the corroboration correct because the topic discussion, ontology link to
VTEB, and eligible cash-flow signal aligned. Annotator~B judged it incorrect
because the transcript did not directly mention VTEB. The annotators
revisited the evidence and Rule~4 criteria. Because Rule~4 permits
corroboration through a topic--product association, a direct product mention
was not required. The relationship was therefore resolved as correct.
This case illustrates how the rule criteria guided resolution of a disputed
judgment.
\end{multicols}
\endgroup

\end{document}